%% file: acl_latex.tex
\pdfoutput=1

\documentclass[11pt]{article}

\usepackage[final]{acl}

\usepackage{times}
\usepackage{latexsym}

\usepackage[T1]{fontenc}

\usepackage[utf8]{inputenc}

\usepackage{microtype}

\usepackage{inconsolata}

\usepackage{graphicx}
\usepackage{booktabs}
\usepackage{float} 
\usepackage{amsmath}

\usepackage{enumitem}

\usepackage{multirow}
\usepackage{longtable}
\usepackage{graphicx}
\usepackage[table]{xcolor}
\usepackage{tabularx}
\usepackage{listings}
\usepackage{xcolor}
\usepackage{placeins}
\usepackage{needspace}
\usepackage{diagbox}

\title{\demoname: Bridging Registries and Literature for Comprehensive Clinical Trial Access}

\author{\textbf{Jiwoo Park} \quad \textbf{Ruoqi Liu} \quad \textbf{Avani Jagdale} \quad \textbf{Andrew Srisuwananukorn} \\ [3pt] \quad \textbf{Jing Zhao} \quad \textbf{Lang Li} \quad \textbf{Ping Zhang} \quad \textbf{Sachin Kumar} \\ [8pt]
The Ohio State University \\
\texttt{\small park.3620@osu.edu}
}
\usepackage{xspace}
\newcommand{\demoname}{\textsc{ClinicalTrialsHub}\xspace}

\newcommand{\Sref}[1]{\S\ref{#1}}

\begin{document}
\maketitle
\begin{abstract}
We present \demoname, an interactive search-focused platform that consolidates all data from ClinicalTrials.gov and augments it by automatically extracting and structuring trial-relevant information from PubMed research articles. Our system effectively increases access to structured clinical trial data by 83.8\% compared to relying on ClinicalTrials.gov alone, with potential to make access easier for patients, clinicians, researchers, and policymakers, advancing evidence-based medicine. \demoname uses large language models such as GPT-5.1 and Gemini-3-Pro to enhance accessibility. The platform automatically parses full-text research articles to extract structured trial information, translates user queries into structured database searches, and provides an attributed question-answering system that generates evidence-grounded answers linked to specific source sentences. We demonstrate its utility through (1) a user study involving clinicians, clinical researchers, and PhD students of pharmaceutical sciences and nursing, and (2) a systematic automatic evaluation of its information extraction and question answering capabilities.\footnote{A demonstration video is available at: \url{https://www.youtube.com/watch?v=uCPxyw7Abh0}. The source code to run the demo locally is available at: \url{https://github.com/jiwoo-jus/clinical-trials-hub}. Evaluation code for comparing model performance is available at: \url{https://github.com/jiwoo-jus/clinical-trials-hub-evaluation}.}
\end{abstract}

\section{Introduction}
Access to clinical trial information is essential for patients seeking new treatments and for clinicians, researchers, and policymakers working to advance medical care. Containing over 500K registrations, \href{https://www.ClinicalTrials.gov}{ClinicalTrials.gov} (CTG) is the primary resource many of these groups use to search and identify ongoing and completed trials. All trials in CTG are available in a structured format allowing the users to easily navigate, filter, or download them. However, many trials remain unregistered or are reported only in publications, particularly those conducted outside the US. \href{https://pubmed.ncbi.nlm.nih.gov/}{PubMed}, which indexes over 35 million biomedical papers and abstracts, often includes results from these unregistered trials. As shown in \autoref{tab:coverage}, a substantial number of trials accessible from PubMed are not registered on CTG. %
The difference in coverage reflects the lack of integration between trial registries and published literature. Despite both being critical resources, PubMed and CTG exist as isolated silos with incompatible formats. PubMed, containing free-text articles, can be especially difficult for non-researchers to parse or filter, requiring substantial manual effort that could take weeks or months.

\begin{table}
\centering
\resizebox{\columnwidth}{!}{
\begin{tabular}{@{}lll@{}}
\toprule
\textbf{Database} & \textbf{Filter/Status} & \textbf{Count} \\
\midrule
CTG & All Studies & 543,172 \\
CTG & Completed w/ Results & 60,449 \\
\midrule
PubMed & Max Sensitivity$^a$ & 962,774 \\
PubMed & Max Specificity$^a$ & 138,279 \\
PubMed & Max Sensitivity + CTG linked$^b$ & 63,928 \\
PubMed & Max Specificity + CTG linked$^b$ & 35,671 \\
\bottomrule
\end{tabular}
}
\caption{Clinical trial data coverage comparison. 
$^a$Identified using PubMed's Clinical Queries search strategies based on~\citep{wilczynski2011sensitive} with PMC Open Access restriction. \textit{Max Sensitivity} prioritizes recall; \textit{Max Specificity} prioritizes precision.
$^b$Subset already registered on ClinicalTrials.gov.}
\label{tab:coverage}
\end{table}

To address this gap, we present \demoname, a unified user-centered trial search platform that combines trial registry data from CTG and published literature from PubMed into a single interface (\Sref{sec:ui}). Our system makes three main contributions. First, our interface accepts natural language queries and displays unified search results combining trials from both sources in a single ranked list, merging overlapping trials (\Sref{sec:search}). Second, using large language models (LLMs), we extract \textbf{structured information} from free text PubMed articles making up to 899,846 previously unregistered clinical trials easily searchable and filterable (\Sref{sec:ie}). 
Lastly, we build an attributed \textbf{question answering} feature that allows the users to ask questions over individual trials (\Sref{sec:qa}); answers are generated with attribution provided in the trial text. 

Our system showcases practical use cases spanning diverse medical contexts, from evidence curation and systematic reviews to dosing protocols and treatment evaluation. We validate the system's utility through comprehensive evaluations comparing frontier LLMs (Gemini-3-Pro \citep{google2025gemini3_pro_system_card}, GPT-5.1 \citep{openai2025gpt5_1_system_card}, Claude-4.5-Sonnet \citep{anthropic2025claude4_5_system_card}) to select optimal models for each feature. We conduct quantitative benchmarking to assess information extraction accuracy using a curated benchmark and evaluate question-answering quality using the FACTS grounding benchmark \citep{jacovi2025factsgroundingleaderboardbenchmarking}. Additionally, we perform a user study with seven medical professionals. The evaluations confirm the platform's effectiveness across its core functionalities and showcase its utility over using PubMed or CTG alone.

\section{\demoname UI Design}
\label{sec:ui}
This section describes the user interface of our system. The platform consists of two primary pages: a main search page and individual trial detail pages.

\begin{figure*}[t]
  \centering
  \includegraphics[width=\textwidth]{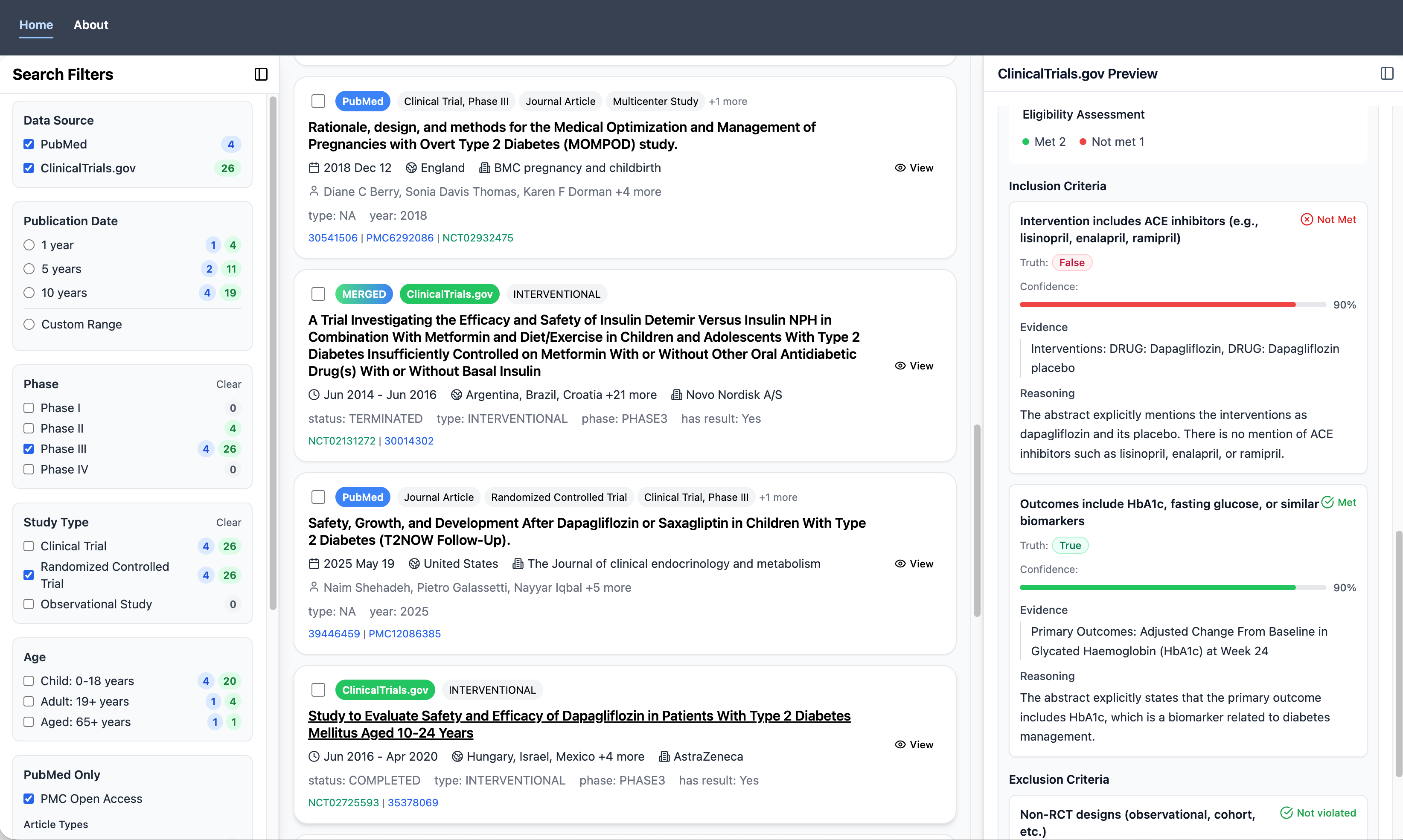}
  \caption{Main search interface with unified results, filters, and eligibility preview}
  \label{fig:SearchPage}
\end{figure*}

\paragraph{Main Search Page.} The main search interface (an overview is shown in \autoref{fig:SearchPage}; additional screenshots are available in \autoref{app:UI-SearchPage}) offers a unified search experience. In \textbf{search bar panel}, users can enter natural language queries (top) or use structured input forms specifying condition, intervention, and other terms for PICO-based search (middle), or utilize expert-level PubMed/CTG query forms (bottom). The \textbf{search results} display a ranked list of trials from both CTG and PubMed sources, with merged entries when a trial appears in both databases. A \textbf{filtering sidebar} enables refinement by data source, study type, phase, design allocation, and other trial-specific categories. The \textbf{details sidebar} provides a quick preview of abstracts and metadata when clicking trial titles. \demoname also provides two experimental features to enhance the search experience. Users can specify \textbf{inclusion and exclusion criteria} for early-stage screening; when viewing a trial's preview sidebar, the system provides a quick eligibility assessment based on the abstract and metadata. Additionally, an \textbf{AI-insight} feature generates five context-aware insights based on the current search results, supporting iterative Q\&A by maintaining context from the user's queries and recent interactions.

\begin{figure*}[t]
  \centering
  \includegraphics[width=\textwidth]{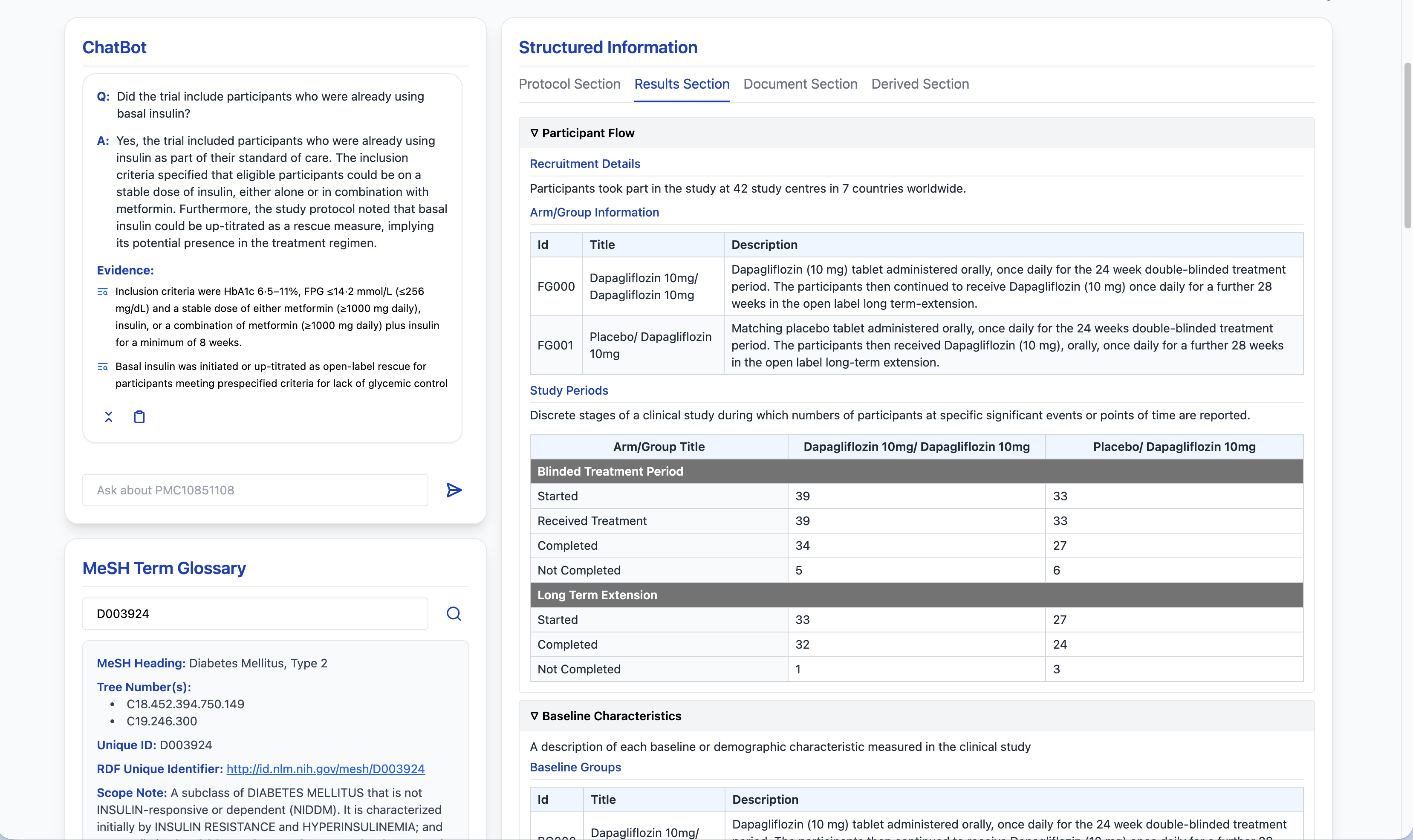}
  \caption{Trial detail page with structured information and evidence-grounded QA}
  \label{fig:DetailPage}
\end{figure*}

\paragraph{Individual Trial Detail Page.} Upon selecting a trial, users access a comprehensive detail page (an overview is shown in \autoref{fig:DetailPage}; additional screenshots are provided in \autoref{app:UI-DetailPage}). Each entry is categorized as \textit{CTG-only}, \textit{PubMed-only}, or \textit{merged} when bidirectional references link a CTG registration to PMC articles. The detail page provides three main components: \textbf{Structured Information} displays trial metadata in an easy-to-navigate format, directly available for CTG entries and extracted from free text for PubMed-only items; \textbf{Full Text} shows the publication text for PubMed entries, while for CTG entries referenced PMC articles can be expanded inline. Since mappings are not always one-to-one---interim reports or sub-studies may reference a single registration, or multiple registrations across trial phases may cite the same final article---the detail page surfaces all linked records and allows toggling between them; information extraction is performed independently for each article. \textbf{Interactive QA} generates evidence-grounded responses linked to specific source sentences that auto-scroll and highlight upon clicking a citation.

\section{\demoname Search}
\label{sec:search}

\begin{figure}[htbp]
    \centering
    \includegraphics[width=0.8\linewidth]{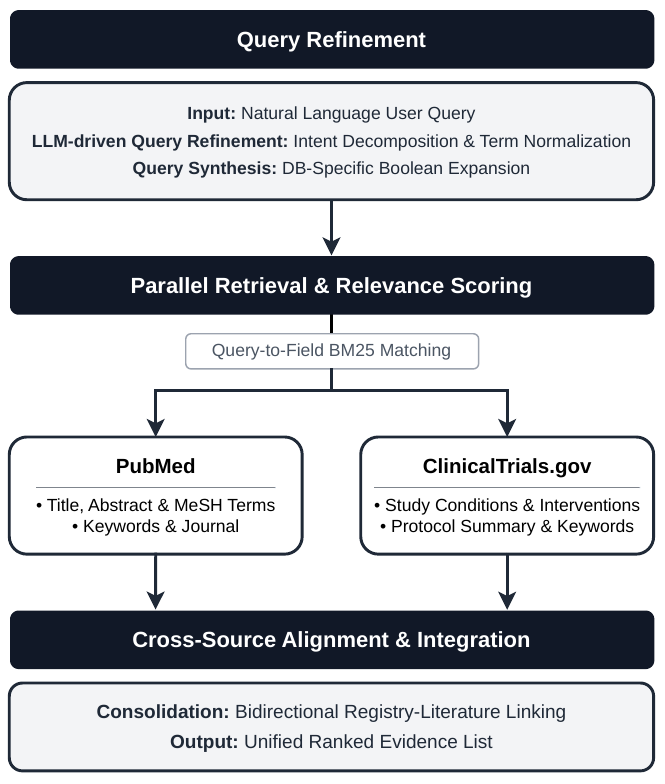}
    \caption{Search pipeline}
    \label{fig:search_pipeline}
\end{figure}

Our search system comprises three components: (1) query refinement that converts natural language queries to platform-specific structured queries, (2) multi-source search on the two platforms, (3) relevance reranking and deduplication. \autoref{fig:search_pipeline} illustrates the overall pipeline.

\subsection{Query Refinement}

The query refinement component translates user queries into structured search parameters.
Our implementation handles both structured field inputs and natural language queries. When users provide individual field values (condition, intervention, other terms), we use them directly. For natural language queries, we use an LLM (GPT-5.1) that applies clinical terminology normalization to decompose the required fields.
We design prompt templates to extract condition terms, intervention specifications, and auxiliary keywords. The prompt used to extract these meanings from user queries is shown in \autoref{app:QueryGeneration-Prompt}.

The extracted output maps directly to database-specific query construction, where PubMed searches combine the structured parameters with Boolean operators and predefined clinical trial detection patterns, while CTG requires the decomposed fields as separate API parameters. This approach maintains consistent query semantics across heterogeneous database interfaces while preserving the original search intent.

\subsection{Multi-Source Search}

Both CTG and PubMed offer their own search APIs. Given the refined query, we execute both APIs asynchronously with platform-specific optimization strategies. PubMed retrieval uses NCBI's \href{https://www.ncbi.nlm.nih.gov/books/NBK25499/}{E-utilities API} in a two-stage process (E-SEARCH for Boolean query execution and PMID extraction, followed by E-FETCH for batch metadata retrieval including abstracts, MeSH terms, and cross-references). CTG integration combines the official REST API for basic trial information with the daily-updated AACT PostgreSQL mirror for enhanced metadata and complex relational queries.

\subsection{Reranking and Result Integration}

We apply BM25 scoring \citep{robertson1994some} separately to each source, normalize scores, and add bonuses based on the similarity ranking within each source. We bidirectionally merge PubMed and CTG entries that reference each other, providing a relevance bonus to merged pairs. All entries are then sorted by final scores and displayed to users. Complete technical details including corpus construction, scoring formulas, and deduplication algorithms are provided in \autoref{sec:Search-BM25}.

\section{Information Extraction}
\label{sec:ie}
All entries on CTG follow a structured format that allows users to navigate and filter across different trials as well as download them for analyses. For example, a patient looking for phase 3 trials for their gender and age range can easily apply a filter and find relevant trials. On the other hand, PubMed articles are research papers in free text form which makes it harder to perform such tasks. Hence, we implement an information extraction pipeline that extracts all relevant trial-related data from papers in the same format as CTG, making indexing, filtering, browsing, and analysis through them easier. We employ an LLM for this purpose that takes as input (parts of) the paper and generates the structured output in a JSON format. 

We perform information extraction for \textbf{211} fields in parallel by dividing the task into six modules for the protocol section, four modules for the results section, and one module for derived section, each with explicit field definitions and enumerated value constraints constructed based on CTG data schema\footnote{\url{https://beta-ut.clinicaltrials.gov/api/v2/studies/metadata}} fed to the LLM (see fields list at \autoref{tab:ie-demo-fields} in \autoref{sec:app-ie}). This approach minimizes both user wait times and hallucinations caused by the LLM processing overwhelming amounts of information. We describe the modules in detail in \autoref{sec:app-ie}, which also includes the prompt templates (\autoref{app:IE-Prompt}).

\subsection{Evaluation}
We evaluated three frontier LLMs (Gemini-3-Pro, GPT-5.1, and Claude-4.5-Sonnet) to measure structural completeness (key-level: whether the field exists) and semantic accuracy (value-level: whether the extracted value matches the ground truth). 

\paragraph{Dataset construction}
We create an evaluation dataset comprising 100 PMC-CTG trial pairs. To do this, we queried PubMed for clinical trials (\autoref{app:sample-id-filter-query-pm}) published after 2021 (the current data schema's modernization point) with PMC full-text availability and CTG status set to completed. We then check for a one-to-one mapping verified bidirectionally through the PMC and CTG databases. We focus on the \textbf{21} distinct field types in protocol section covering study design, arm group interventions, outcome measures, and eligibility criteria. 

We excluded CTG fields appearing in fewer than 15\% of records and fields primarily found in PMC articles but inconsistently updated on CTG, considering them unsuitable as reliable ground truth. Since results reporting has been required, limited enforcement has led to results information being more complete in PMC articles, while CTG entries may be missing or outdated. Conversely, CTG often lists extensive secondary outcomes and adverse events, whereas PMC articles emphasize primary outcomes. We plan to create a systematically validated benchmark dataset covering this results section as our future work. Model predictions and ground truth data are available \href{https://drive.google.com/drive/folders/1BwghEAHvYblMgT9pffi-5xWlASmftAWL?usp=sharing}{here}.

For key-level evaluation, we flatten all fields from both model prediction and reference, preserving hierarchical relationships. For each field in the schema, we classify its extraction status as: \textit{True Positive} (field exists in both prediction and reference), \textit{False Positive} (field appears in prediction but not defined in the CTG data schema), \textit{False Negative} (field exists in reference but missing from prediction), or \textit{Extra Valid} (field appears in prediction and is valid according to the schema, but absent from this specific reference case, making it unevaluable for correctness).

For value-level evaluation, for fields with \textit{True Positive} status, we assess extracted values using a three-stage approach. First, we perform exact string matching with normalization. For mismatches, we calculate semantic similarity using GPT-5.1 as judge with a 0.70 threshold (prompt shown in \autoref{app:semantic-prompt}). We choose this threshold qualitatively by manually analyzing model outputs. For structured list fields, we create an $m \times n$ similarity matrix where $m$ is the number of reference elements and $n$ is the number of predicted elements to score all reference-prediction element pairs, then use the Hungarian algorithm \citep{kuhn1955hungarian}
to optimize total similarity across matched pairs, allowing accurate per-element evaluation regardless of list order.

\paragraph{Results}
Table \autoref{tab:extraction_metrics} summarizes the overall performance. GPT-5.1 achieved the highest performance across all metrics, recording a Key-level F1 of 0.980 and a Value-level F1 of 0.890. Gemini-3-Pro exhibited competitive performance (Key F1 0.960, Value F1 0.870), whereas Claude-4.5-Sonnet showed a clear performance gap (Key F1 0.740, Value F1 0.660). 
However, we note that the differences between GPT-5.1 and Gemini-3-Pro are small across both key- and value-level metrics. Because we used GPT-5.1 as the semantic similarity judge, a small bias in its favor is possible. To mitigate this, we manually reviewed some of the disagreement cases and confirmed that the judged matches were generally valid.
Given the overall closeness of model performance, our results suggest that either GPT-5.1 or Gemini-3-Pro could serve as a reasonable backbone model in practice. We select GPT-5.1 for our system primarily for cost-efficiency and practical deployment considerations rather than because of any substantial performance gap.

A more fine-grained view is presented in \autoref{tab:ie-eval-fields}, which reports value-level F1 scores for each individual field alongside the average number of Extra-Valid (EV) cases. Although EV cases cannot be directly evaluated due to the lack of a reference value, they nevertheless reveal meaningful patterns about the underlying data sources. Several descriptive and eligibility-related fields exhibit high EV frequencies, indicating that PMC articles often contain information that is absent in CTG. Importantly, these same fields also show high F1 scores when CTG does provide a reference value, suggesting that discrepancies arise primarily from CTG incompleteness rather than model hallucination. This pattern reinforces the complementary value of LLM-based extraction when dealing with fields that are inconsistently maintained or only partially populated in CTG.

\begin{table}[t]
\centering
\small
\resizebox{\columnwidth}{!}{%
\begin{tabular}{llccc}
\toprule
\textbf{Level} & \textbf{Model} & \textbf{Precision} & \textbf{Recall} & \textbf{F1} \\
\midrule
\multirow{3}{*}{Key} 
  & Gemini-3-Pro      & \textbf{1.000} & 0.930 & 0.960 \\
  & GPT-5.1           & \textbf{1.000} & \textbf{0.970} & \textbf{0.980} \\
  & Claude-4.5-Sonnet & 0.900          & 0.690          & 0.740 \\
\cmidrule(lr){1-5}
\multirow{3}{*}{Value}
  & Gemini-3-Pro      & 0.830          & 0.910          & 0.870 \\
  & GPT-5.1           & \textbf{0.840} & \textbf{0.960} & \textbf{0.890} \\
  & Claude-4.5-Sonnet & 0.740          & 0.670          & 0.660 \\
\bottomrule
\end{tabular}%
}
\caption{Information extraction performance across key-level and value-level metrics.}
\label{tab:extraction_metrics}
\end{table}

\section{Interactive Grounded QA}
\label{sec:qa}
On the trial details page, we provide an LLM-powered chat interface that allows the user to ask question about individual trials with evidence-grounded responses. To select the optimal model for this component, we evaluated three frontier LLMs (Gemini-3-Pro, GPT-5.1, Claude-4.5-Sonnet).
\paragraph{Benchmark Selection}
For this evaluation, we used the FACTS Grounding benchmark  \citep{jacovi2025factsgroundingleaderboardbenchmarking}
, which measures a model's ability to generate responses factually grounded in provided context documents. The benchmark aligns directly with our requirements by incorporating substantial medical-domain coverage (29\%), supporting long context documents up to 32K tokens with diverse query patterns, and explicitly constraining models to rely solely on the provided context.
\paragraph{Evaluation Methodology}
FACTS employs a two-phase evaluation conducted by three judge models (Gemini-1.5-Pro, GPT-4o, and Claude-3.5-Sonnet).
\textbf{1. Instruction Following (Eligibility):} Judges assess whether responses adequately address the user request, assigning verdicts of \textit{No issues,} \textit{Minor issue(s),} or \textit{Major issue(s)}. A response is deemed ineligible (and assigned a score of 0) if all three judges classify it as having major issues, preventing it from advancing to the grounding evaluation phase.
\textbf{2. Grounding}: Judges evaluate whether each response is fully grounded in the source document by classifying every sentence as \textit{supported}, \textit{unsupported}, \textit{contradictory}, or \textit{no\_rad} (requires no factual grounding). A response receives a positive grounding verdict from a judge only if all sentences are either supported or \textit{no\_rad}. The factuality score for each response is the average of the three judges' verdicts, and each model's final score is the mean factuality score across all evaluated responses.

\paragraph{Evaluation setup}
We evaluated Gemini-3-Pro, GPT-5.1, and Claude-4.5-Sonnet on 236 medical-domain samples from FACTS, generating all responses but evaluating the top 100 shortest average response length to balance cost and significance following FACTS guideline. 

For grounding evaluation, we used the \textit{JSON} prompt template from the available templates, which FACTS identified as optimal for Gemini-1.5-Pro and GPT-4o. This JSON template instructs the model to output structured JSON objects for each sentence, with classification labels, explicit rationales explaining each decision, and supporting evidence excerpts from the context document. 

In contrast, the \textit{implicit-span-level} prompt template generates unstructured natural language output, listing each sentence with a simple binary accurate/inaccurate label and concluding with a final verdict, without providing detailed rationales or supporting evidence. In FACTS, this approach had previously been preferred for Claude-3.5-Sonnet due to its lower complexity in structured output generation. However, Claude-4.5-Sonnet exhibited substantially improved JSON generation reliability, allowing us to use the more informative template format (see \autoref{tab:facts-judge-prompts}).

\paragraph{Results}
Gemini-3-Pro achieved the highest grounding score (0.897), notably outperforming GPT-5.1 (0.680) and Claude-4.5-Sonnet (0.767), with consistently stronger performance across all three evaluators (see \autoref{tab:QA-results}). We additionally recorded completion tokens, response time, and response length for all generated responses. As shown in \autoref{tab:model_runtime_stats}, while Gemini-3-Pro required significantly higher completion tokens (1,318 avg. vs. $\sim$220 for others) and generation time (14.33s vs. $\sim$4s), it produced the most concise final answers (698 characters vs. $\sim$950). This suggests that Gemini performs a deeper internal reasoning process to synthesize grounded, compact responses; this behavior is critical for trustworthy clinical QA. Consequently, we selected Gemini-3-Pro as our QA backbone. 

\begin{table}[t]
\centering
\small
\resizebox{\columnwidth}{!}{%
\begin{tabular}{lccccc}
\toprule
\diagbox{\textbf{Model}}{\textbf{Judge}} & \textbf{Average} & \textbf{Gemini} & \textbf{GPT} & \textbf{Claude} \\
\midrule
Gemini-3-Pro      & \textbf{0.897} & \textbf{0.920} & \textbf{0.830} & \textbf{0.930} \\
GPT-5.1           & 0.680          & 0.670          & 0.610          & 0.760 \\
Claude-4.5-Sonnet & 0.767          & 0.740          & 0.690          & 0.860 \\
\bottomrule
\end{tabular}%
}
\caption{Evaluation scores by judge model. Each row shows a prediction model evaluated by three different judge models (Gemini-3-Pro, GPT-5.1, Claude-4.5-Sonnet) and their average score.}
\label{tab:QA-results}
\end{table}

\begin{table}[t]
\centering
\small
\resizebox{\columnwidth}{!}{%
\begin{tabular}{lrrr}
\toprule
\textbf{Model} & \textbf{Tokens} & \textbf{Time} & \textbf{Length} \\
\midrule
Gemini-3-Pro      & 1318.44 & 14.33 & 698.01 \\
GPT-5.1           & 227.71  & 2.81  & 1006.82 \\
Claude-4.5-Sonnet & 210.95  & 5.34  & 896.54 \\
\bottomrule
\end{tabular}%
}
\caption{Performance comparison. 
\textbf{Tokens}: Avg. completion tokens; 
\textbf{Time}: Avg. latency (s); 
\textbf{Length}: Avg. output characters.}
\label{tab:model_runtime_stats}
\end{table}

\section{User Study}
\label{sec:user_study}
To assess the practical utility of \demoname, we conducted an initial user study with seven medical professionals: hematologists, pathologists, dentists, clinical statisticians, pharmacists, pharmaceutical scientists, and nursing researchers. Participants explored the system for tasks corresponding to their typical use of PM and CTG. \autoref{tab:user_study_results} summarizes key findings. Detailed study materials are provided in \autoref{sec:user-study}.

\begin{table}[h]
\centering
\small
\begin{tabularx}{\linewidth}{@{}Xc@{}} 
\toprule
\textbf{Metric} & \textbf{Score} \\
\midrule
\multicolumn{2}{@{}l}{\textit{Search Feature Experience (0--5):}} \\
Query generation rating & 4.50 \\
Filtering rating & 4.40 \\
Eligibility check rating & 4.50 \\
Combined search utility & 4.14 \\
\midrule
\multicolumn{2}{@{}l}{\textit{Information Extraction Accuracy (0--5):}} \\
Study Overview & 5.0 \\
Study Plan & 5.0 \\
Participation Requirements & 4.83 \\
Baseline Characteristics & 4.83 \\
Outcome Measures & 4.83 \\
Participant Flow & 4.7 \\
Adverse Events & 4.6 \\
\midrule
Chatbot answer quality \textit{(0--5)} & 4.86 \\
\bottomrule
\end{tabularx}
\caption{User study results summary. Scores reflect participant ratings of ClinicalTrialsHub functionality.}
\label{tab:user_study_results}
\end{table}
\textbf{Search Stage.} Six of seven participants found 6 or more relevant studies with \demoname compared to 5/7 for PubMed and 3/7 for CTG among the top 30 results, demonstrating improved relevance through unified search with BM25 reranking. Query generation (4.50), filtering (4.40), and eligibility checking (4.50) features received high ratings. The overall combined search capability received strong approval (4.14), validating the value of eliminating cross-platform navigation. Time-saving received moderate rating (3.71), potentially reflecting initial learning curve.

\textbf{Review Stage.} Six of seven participants reviewed the structured extraction from PubMed. Perceived accuracy was consistently high: Study Overview and Study Plan received perfect scores (5.0), while Participation Requirements, Baseline Characteristics, and Outcome Measures scored 4.83. Results-oriented sections (Participant Flow: 4.7, Adverse Events: 4.6) also rated highly. All seven participants used the chatbot, rating answer quality (4.86) and overall detail page efficiency (4.86) very highly.

\section{Related Work}
Prior work has aimed to improve clinical trial information access and evidence synthesis in various ways.
Trialstreamer \citep{marshall2020trialstreamer} structures PubMed articles for 
rapid evidence browsing. 
RobotReviewer \citep{marshall2015robotreviewer} automates trial data extraction and risk-of-bias evaluation specifically for systematic reviews. 
LinkedCT \citep{hassanzadeh2009linkedct} transforms ClinicalTrials.gov data into structured linked data though it does not integrate literature sources, unlike our work. Prior work studied the use of LLMs this space as well---for patient-trial matching, clinical trial design, and participant recruitment~\citep{wang2024acceleratingclinicalevidencesynthesis}. Most recently, TrialPanorama \citep{wang2025trialpanoramadatabasebenchmarksystematic} established a large-scale database and benchmark for these tasks, while LEADS \citep{wang2025leads} introduced a foundation model specifically designed to enhance human-AI collaboration in medical literature mining.
\demoname extends these approaches into a unified platform, bridging registry and literature silos. Unlike Trialstreamer or LinkedCT, it integrates both data sources, and compared to RobotReviewer’s narrow focus on systematic reviews, it supports broader interactive exploration and structured retrieval for diverse clinical and research tasks.

\section{Conclusion and Future Work}
We presented \demoname, a unified platform that integrates structured trial registry data from ClinicalTrials.gov with structured information extracted from unstructured PubMed publications using LLMs. Our system enhances access to comprehensive clinical trial information by enabling unified search, structured information extraction and attributed question answering, supporting the diverse needs of patients, clinicians, and researchers.

Several directions remain for future work. Our extraction evaluation currently covers 21 protocol-level fields; extending this to results-related fields with validated benchmarks is an important next step. We also plan to systematically compare our BM25-based retrieval with dense retrieval methods, conduct larger-scale user studies, and improve extraction accuracy through domain-adapted fine-tuning and error analysis across field types.

\newpage
\bibliography{custom}
\appendix
\section{Additional Statistics Details}
\label{sec:appendix}
Our data expansion is calculated as:
\begin{equation}
\text{Expansion(\%)} = \frac{\text{PMC trials without CTG}}{\text{Total PMC trials}} \times 100
\end{equation}
This yields expansion rates of 74.2\% (specificity: 102,608/138,279) to 93.4\% (sensitivity: 898,846/962,774), averaging 83.8\%.
\FloatBarrier
\section{Reranking and Result Integration Details}
\label{sec:Search-BM25}
We implement BM25-based reranking mechanisms \citep{robertson1994some} to individual search results, followed by deduplication to merge related publications and trials.
\paragraph{Corpus Construction:} For PubMed documents, we construct searchable text by concatenating: (1) article titles, (2) abstracts (structured or unstructured), (3) author-provided keywords, (4) Medical Subject Headings (MeSH) descriptors, and (5) journal names. For CTG documents, we concatenate: (1) trial titles, (2) condition specifications, (3) brief summaries, and (4) keywords.
\paragraph{BM25 Scoring:} During integration, BM25 scores are computed for each source separately, then min-max normalized to the $[0,1]$ range and combined with a position-based bonus from the original API ranking. For each document $d_i$, the score is computed as:
\begin{equation*}
\frac{\text{BM25}(d_i) - \text{BM25}_{\min}}{\text{BM25}_{\max} - \text{BM25}_{\min}} + 0.2 \cdot \frac{N - i}{N}
\end{equation*}
where $\text{BM25}(d_i)$ represents the raw BM25 score, $\text{BM25}_{\min}$ and $\text{BM25}_{\max}$ are the minimum and maximum scores for normalization, $N$ is the total number of results, and $i$ is the zero-indexed position in the original API ranking. The constant $0.2$ represents the maximum position bonus.
When $\text{BM25}_{\max} = \text{BM25}_{\min}$ (indicating identical scores), the BM25 component is set to 0.0 for all documents, and the final ranking relies entirely on the original API position.
\paragraph{Bidirectional Deduplication:} We merge PubMed and CTG entries that reference each other.\footnote{This merging criterion is designed for high precision. We acknowledge that might miss certain trials that may have unidirectional references but do not reference each other (for instance, trials that were registered before the publication but never updated)}
Merged pairs receive a relevance bonus.\footnote{We empirically determine this bonus to 0.3.}
After rescoring and merging, all entries are sorted by their final scores and shown to the user.
\section{Information Extraction Details}
\label{sec:app-ie}
\subsection{Modular Prompt System}
The extraction framework is organized into 3 main sections corresponding to the CTG API v2.0 structure. We provide the exact prompts in our repository.
\begin{itemize}[leftmargin=3mm,topsep=0mm,itemsep=0mm]
\item \textbf{protocolSection:} Contains basic study information including identification modules (NCT ID, title, sponsors), status modules (enrollment, dates, phases), design modules (study type, allocation, masking), arms and interventions, eligibility criteria, contacts, and locations.
\item \textbf{resultsSection:} Encompasses study outcomes and results when available, including participant flow, baseline characteristics, outcome measures, adverse events, and statistical analyses.
\item \textbf{derivedSection:} Includes system-generated data elements such as condition browse modules, intervention browse modules, and MeSH term mappings.
\end{itemize}
\subsection{Data Validation and Schema Conformance}
During extraction, we asynchronously operate a validation pipeline.
\begin{itemize}[leftmargin=3mm,topsep=0mm,itemsep=0mm]
\item \textbf{Schema-level validation} enforces structural integrity across all five data sections. Each extracted field undergoes type checking against CTG's built-in types, including character limits for text fields (briefTitle: 300 chars, officialTitle: 600 chars, eligibilityCriteria: 20,000 chars), ISO 8601 date formatting, and proper nesting for complex structures like baseline characteristics and outcome measures.
\item \textbf{Enumerated value validation} ensures all categorical fields contain only permitted values. The system validates against ClinicalTrials.gov's comprehensive enum definitions for critical fields including studyType, allocation, interventionModel, phases, sex, and standardized age groups. Non-conforming values trigger immediate correction through fuzzy matching against allowed values.
\item \textbf{Clinical terminology verification} integrates with the NCBI MeSH API to validate medical subject headings in real-time. The system performs fuzzy matching for condition and intervention terms, automatically suggesting and applying standardized MeSH descriptors. This prevents terminology drift and ensures compatibility with biomedical databases.
\item \textbf{Statistical consistency checking} validates the coherence of quantitative data, verifying group-measure associations in baseline characteristics, confirming unit consistency across measurements, and validating statistical parameters (means, standard deviations, confidence intervals) for mathematical soundness.
\end{itemize}
Missing information is handled through deliberate omission rather than placeholder generation, preserving data integrity for downstream analysis.
\subsection{Sample ID Filter Query - PubMed}
\label{app:sample-id-filter-query-pm}
\begin{lstlisting}[numbers=none,basicstyle=\ttfamily, keywordstyle=, commentstyle=, stringstyle=,showspaces=false,showstringspaces=false]
(("randomized controlled trial"[Publication Type]
 OR "controlled clinical trial"[Publication Type]
 OR "randomized"[Title/Abstract]
 OR "placebo"[Title/Abstract]
 OR "clinical trials as topic"[MeSH Terms:noexp]
 OR "randomly"[Title/Abstract]
 OR "trial"[Title])
 NOT ("animals"[MeSH Terms] NOT "humans"[MeSH Terms]))
 AND ("english"[Language] OR "English"[lang])
 AND "pubmed pmc open access"[Filter]
 AND "clinicaltrials gov"[Secondary Source ID]
 AND "2021/01/01"[Date - Publication] : "3000"[Date - Publication]
\end{lstlisting}
\subsection{Semantic Similarity Prompt}
\label{app:semantic-prompt}
\begin{lstlisting}[numbers=none,basicstyle=\ttfamily, keywordstyle=, commentstyle=, stringstyle=]
Compare these clinical trial field values for '{field_explanation}'. Return only a number 0-1 for semantic similarity.
Text1: {reference_value}
Text2: {predicted_value}
\end{lstlisting}
\section{User Study Details}
\label{sec:user-study}
To assess the practical utility of ClinicalTrialsHub in realistic workflows, we conducted an initial user study with seven medical professionals. Participants were recruited across diverse roles: hematology, pathology, dentistry, clinical statistics, pharmacy, pharmaceutical science (PhD student), and nursing (PhD candidate). The participant pool comprised a balanced distribution of clinicians (3/7), clinical researchers (2/7), and students or trainees (2/7), thereby representing key stakeholder groups who utilize clinical trial data. Each participant received an explanation of the system's features and was asked to use \demoname for tasks that correspond to their typical use of PM and CTG. The contexts in which participants explored baseline systems (PM/CTG) and \demoname, along with their stated purposes, are described in \autoref{tab:US-Scenarios}, while the actual queries they input for each platform are shown in \autoref{tab:US-Search-Queries}.
\subsection{Search Stage Evaluation}
\paragraph{Query Generation}Although only two participants explicitly used the natural language query generation feature, both rated it highly (4.50). This low usage rate is unsurprising given that most participants were already proficient in constructing Boolean queries for PubMed and CTG. \autoref{tab:US-Search-SearchQueryGeneration} shows their natural language inputs and the system-generated structured queries. These results demonstrate that the system accurately decomposes complex natural language requests into structured parameters, particularly benefiting less experienced users or those working outside their primary domain.
\paragraph{Filtering}Five participants used the filtering capabilities and rated this feature highly (4.40). The filters applied spanned temporal constraints (publication date ranges from 5-10 years), study design parameters (study type—interventional/observational, RCT status, data availability requirements (PMC Open Access, CTG with results posted), and population specifications (humans, age restrictions, completion status). \autoref{tab:US-Search-Filters} documents the specific filter combinations each participant employed. The \demoname's filtering mechanism enabled users to impose identical selection criteria on heterogeneous data sources, reducing the need to mentally translate filter semantics between platform-specific interfaces.
\paragraph{Eligibility Check}Four participants utilized the eligibility criteria specification feature, which also received a high average rating (4.50). This feature allowed users to define detailed inclusion and exclusion criteria that went beyond simple filter combinations. For instance, one participant specified inclusion criteria requiring "randomized controlled or single-arm registrational trials with $\ge$ 50 patients" that "reported at least one primary outcome," while explicitly excluding "studies including other myeloproliferative neoplasms without fibrosis." A diabetes researcher defined even more granular criteria spanning \textit{intervention type} ("dyadic or family-based behavioral, psychoeducational, or self-management intervention"), \textit{population characteristics} ("adults $\ge$ 18 years with type 2 diabetes"), \textit{outcome requirements} (diabetes self-efficacy, self-management behaviors, dyadic processes, or HbA1c), and \textit{temporal constraints} ("follow-up of at least 3 months"). The complete set of criteria specifications is provided in \autoref{tab:US-Search-EligibiligyCheck}. The high rating reflects users' need for nuanced eligibility assessment that cannot be captured through simple keyword filtering alone—particularly for systematic review preparation and evidence synthesis where precise population and study design specifications are critical.
\paragraph{Overall System Utility}\autoref{tab:US-relevantStudies} shows the distribution of relevant studies participants identified among the top 30 results. 6 of 7 participants found 6 or more relevant studies with \demoname, compared to 5 of 7 for PubMed and only 3 of 7 for CTG. Although CTH demonstrated improved relevance, the maximum response option of '11+' prevents precise quantification of the magnitude of improvement. Nevertheless, our unified search with BM25 reranking successfully integrated heterogeneous data sources into a single ranking system without degrading user satisfaction. This suggests that reducing manual screening effort through cross-source integration is achievable even when reconciling disparate ranking algorithms.
The overall combined search capability received strong approval (4.14), validating the value proposition of eliminating cross-platform navigation and manual deduplication. However, the time-saving metric received a more moderate rating (3.71). This may reflect the initial learning curve associated with a new interface or, alternatively, that experienced users already possess efficient workflows for rapidly applying trial filters on baseline systems. We view these responses as an opportunity to identify superior interaction patterns from existing systems and either integrate them into ClinicalTrialsHub or better expose our system's capabilities to accelerate clinical trial research activities.
\begin{table}[t]
\centering
\small
\resizebox{\columnwidth}{!}{%
\begin{tabular}{lcccc}
\toprule
\textbf{Platform} & \textbf{0} & \textbf{1--5} & \textbf{6--10} & \textbf{11+} \\
\midrule
PubMed            & 0 & 2 & 2 & 3 \\
ClinicalTrials.gov& 0 & 4 & 2 & 1 \\
ClinicalTrialsHub & 0 & 1 & 3 & 3 \\
\bottomrule
\end{tabular}%
}
\caption{Distribution of relevant studies identified among top 30 results.}
\label{tab:US-relevantStudies}
\end{table}

\subsection{Review Stage Evaluation}
In the review stage, we focused on how well the detail page supported close reading of individual trials. Participants were first asked to rate the accuracy of seven structured sections distilled from PMC (Study Overview, Participation Requirements, Study Plan, Participant Flow, Baseline Characteristics, Outcome Measures, Adverse Events) on a 0–5 scale, then to judge whether this representation and the integrated chatbot detail page helped them interpret studies more efficiently and save time.
\paragraph{Information Extraction}
6 of 7 participants reported reviewing the structured extraction from PubMed within \demoname. Across those respondents, perceived accuracy was consistently high: both the \emph{Study Overview} and \emph{Study Plan} modules received perfect mean scores of 5.0/5.0, while \emph{Participation Requirements}, \emph{Baseline Characteristics}, and \emph{Outcome Measures} clustered tightly of 4.83. Even the more detailed results-oriented sections, \emph{Participant Flow} (4.7) and \emph{Adverse Events} (4.6, with one non-response), were rated near the top of the scale. These scores suggest that, for the protocol-level fields we benchmark in \Sref{sec:ie}, clinicians also subjectively experience the structured representations as faithful to the source.
\paragraph{QA}
All seven participants used the chatbot on the detail page, and they were asked to provide up to three concrete examples (paper ID, question, and answer) from their own sessions. Their questions illustrate how the assistant is used as an interpretive layer rather than a generic Q\&A tool. Several participants asked design and endpoint focused questions, such as summarizing how the sample size was determined, clarifying what the primary endpoint was and whether it was met, or checking whether specific biomarkers  were collected. Others queried safety and practical implications, including requests for the most common adverse events, recommended safest order for clinical steps, or whether particular clinical outcomes showed improvement. Some participants targeted fields in protocol section, such as total enrollment numbers, completion dates, or the geographic distribution of trial sites. Across these diverse uses, participants rated the chatbot’s answers as highly accurate and relevant (4.86) and also agreed that the combined detail page—full text, structured view, and chatbot—helped them interpret study information efficiently (4.86).
\input{tables/US-Scenarios}
\input{tables/US-Search-SearchQueryGeneration}
\input{tables/US-Search-Queries}
\input{tables/US-Search-Filters}
\input{tables/US-Search-EligibiligyCheck}
\input{tables/US-ChatbotInteractions}
\FloatBarrier

\onecolumn
\section{Extraction Schema Fields}
\subsection{Extraction Schema Evaluation Fields}
\input{tables/ie-eval-fields}
\subsection{Extraction Schema All Fields}
\input{tables/ie-demo-fields}
\FloatBarrier
\clearpage
\section{QA Evaluation Details}
\input{tables/facts-judge-prompts}
\FloatBarrier
\clearpage
\section{User Interface - Search Page}
\label{app:UI-SearchPage}
\begin{figure}[H]
  \centering
  \includegraphics[width=\textwidth]{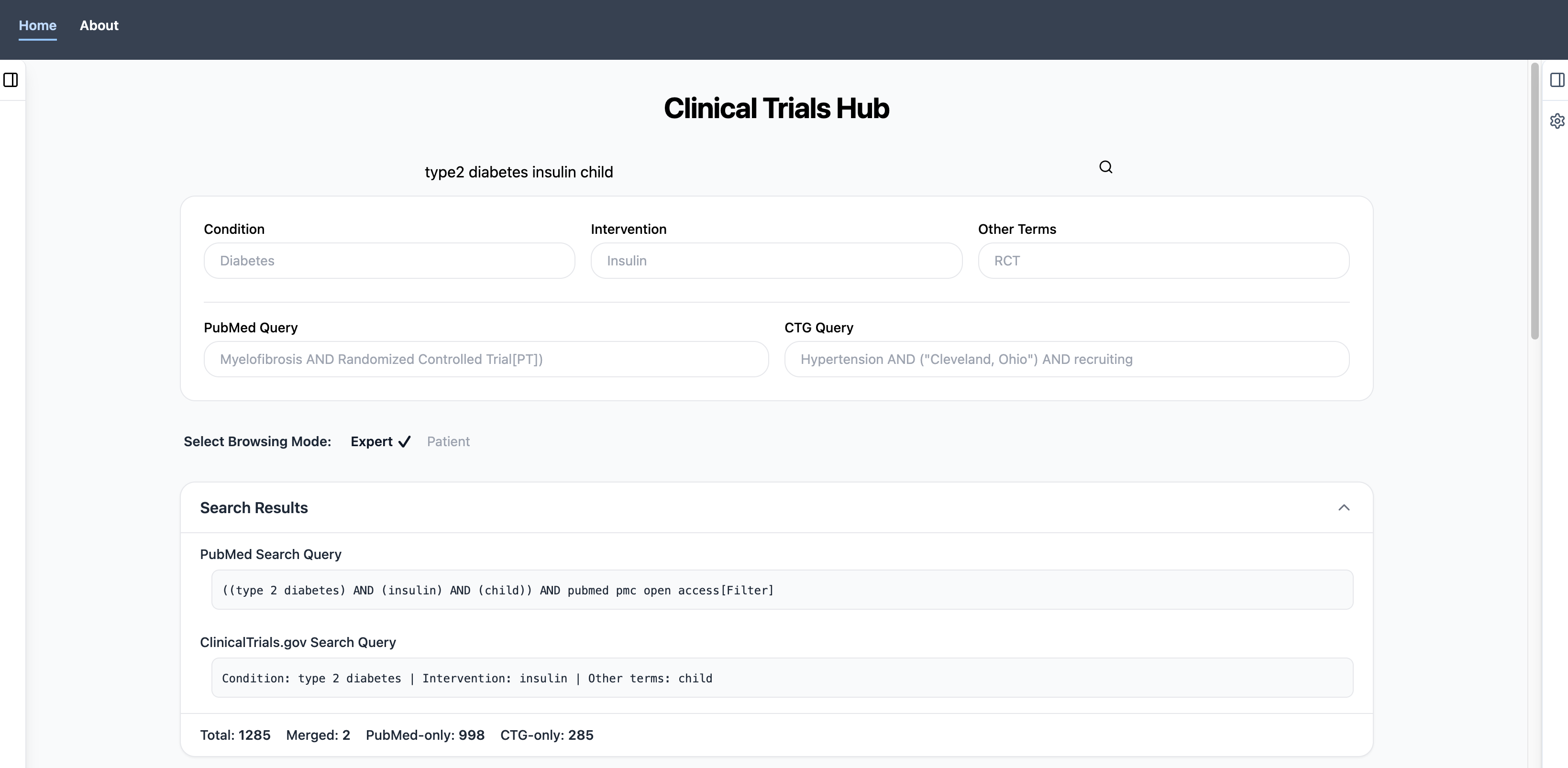}
  \caption{Search panel}
  \label{fig:UI-SearchPage-SearchBar}
\end{figure}
\FloatBarrier
\begin{figure}[H]
  \centering
  \includegraphics[width=\textwidth]{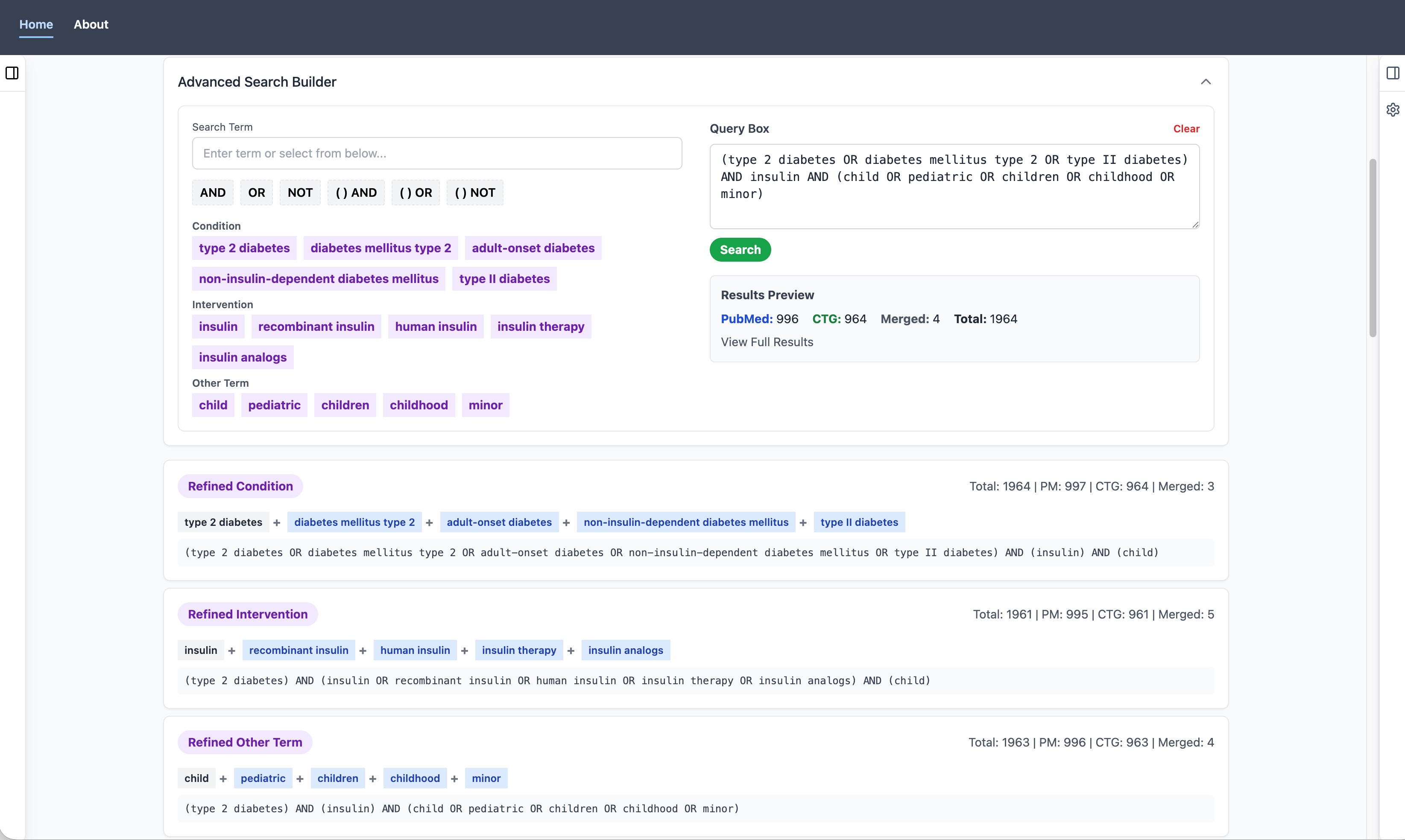}
  \caption{Advanced search panel}
  \label{fig:UI-SearchPage-AdvancedSearch}
\end{figure}
\FloatBarrier
\clearpage
\begin{figure}[H]
  \centering
  \includegraphics[width=\textwidth]{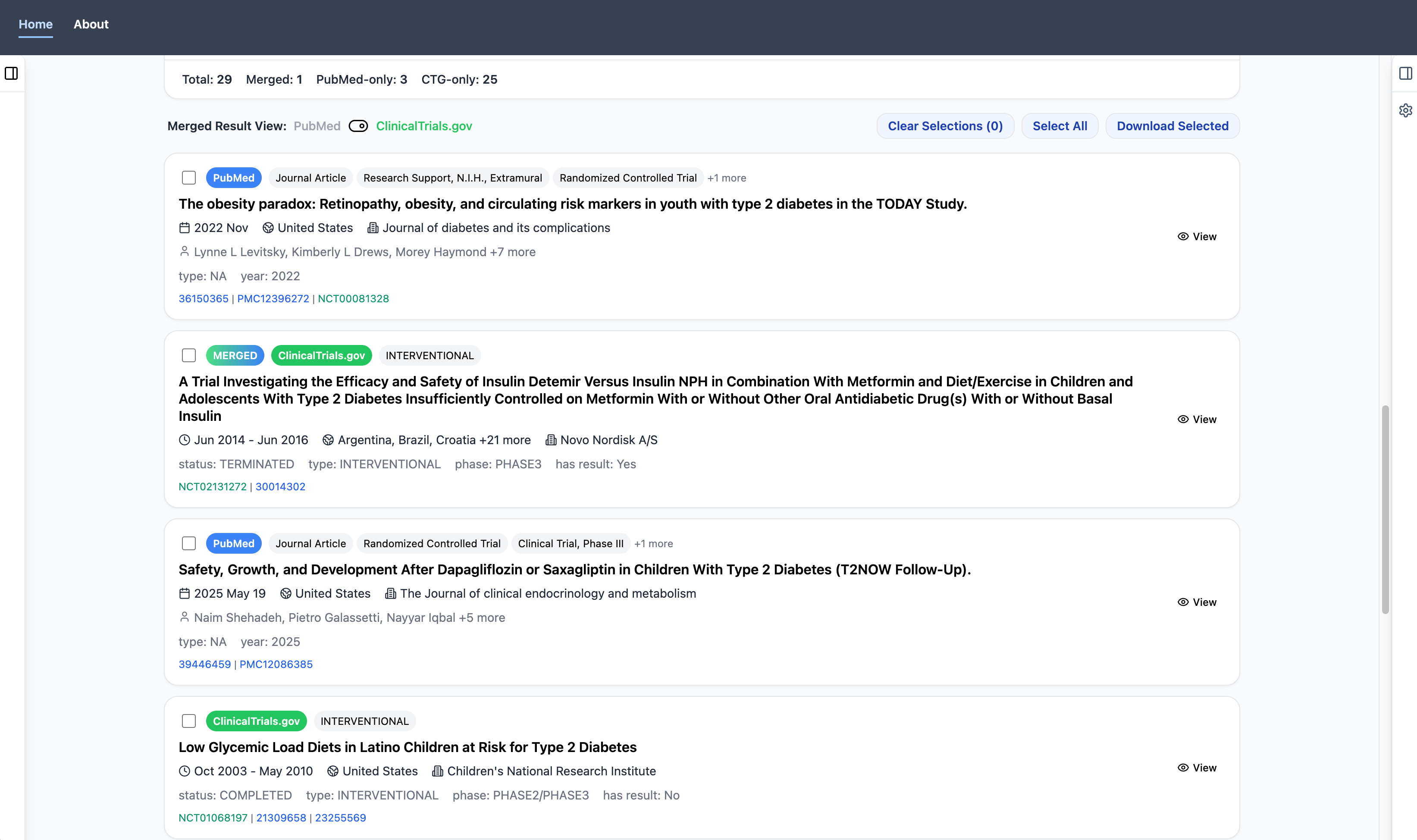}
  \caption{Search results}
  \label{fig:UI-SearchPage-Results}
\end{figure}
\FloatBarrier
\begin{figure}[H]
  \centering
  \includegraphics[width=\textwidth]{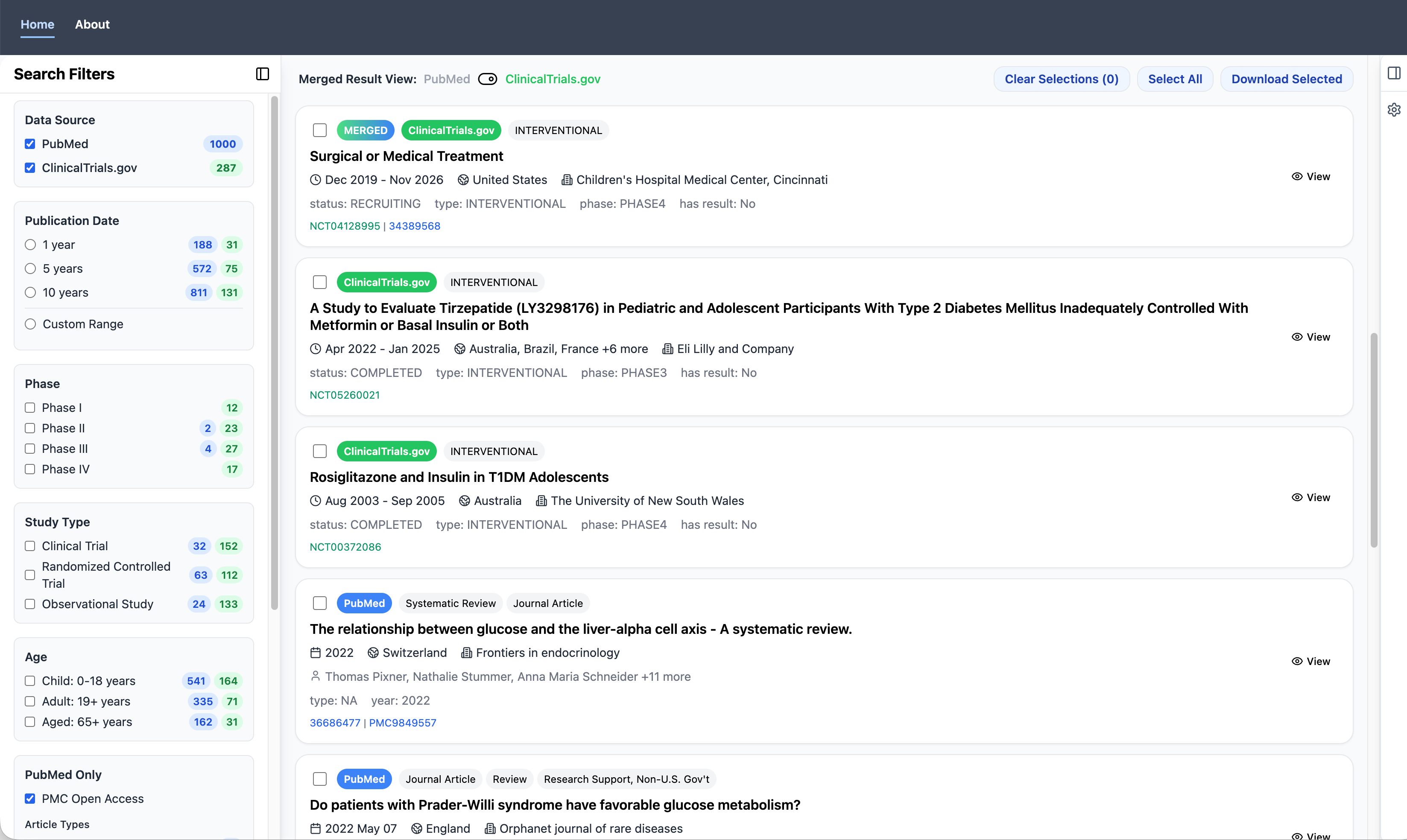}
  \caption{Filtering sidebar}
  \label{fig:UI-SearchPage-Filters}
\end{figure}
\FloatBarrier
\clearpage
\begin{figure}[H]
  \centering
  \includegraphics[width=\textwidth]{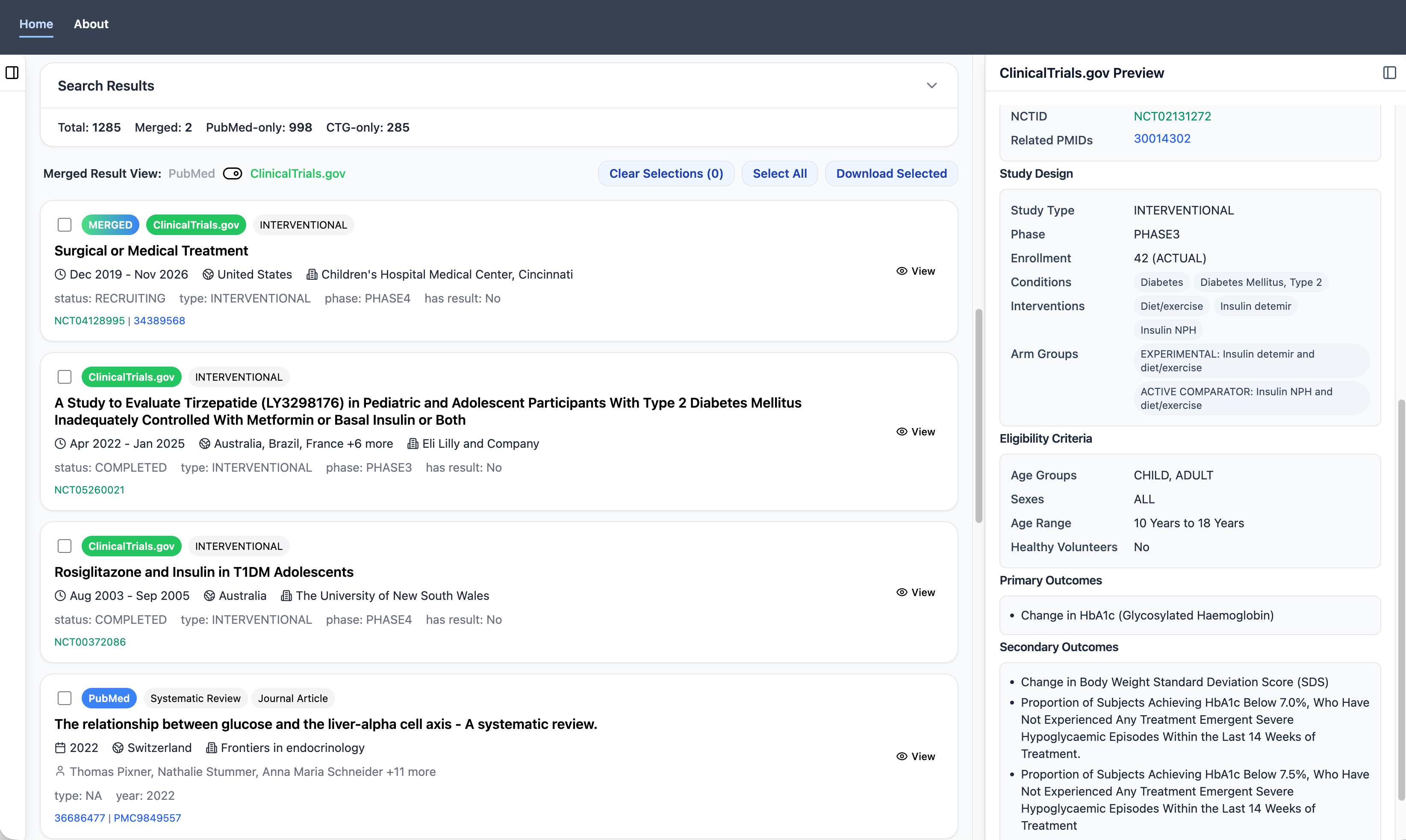}
  \caption{Preview sidebar}
  \label{fig:UI-SearchPage-Preview}
\end{figure}
\FloatBarrier
\begin{figure}[H]
  \centering
  \includegraphics[width=\textwidth]{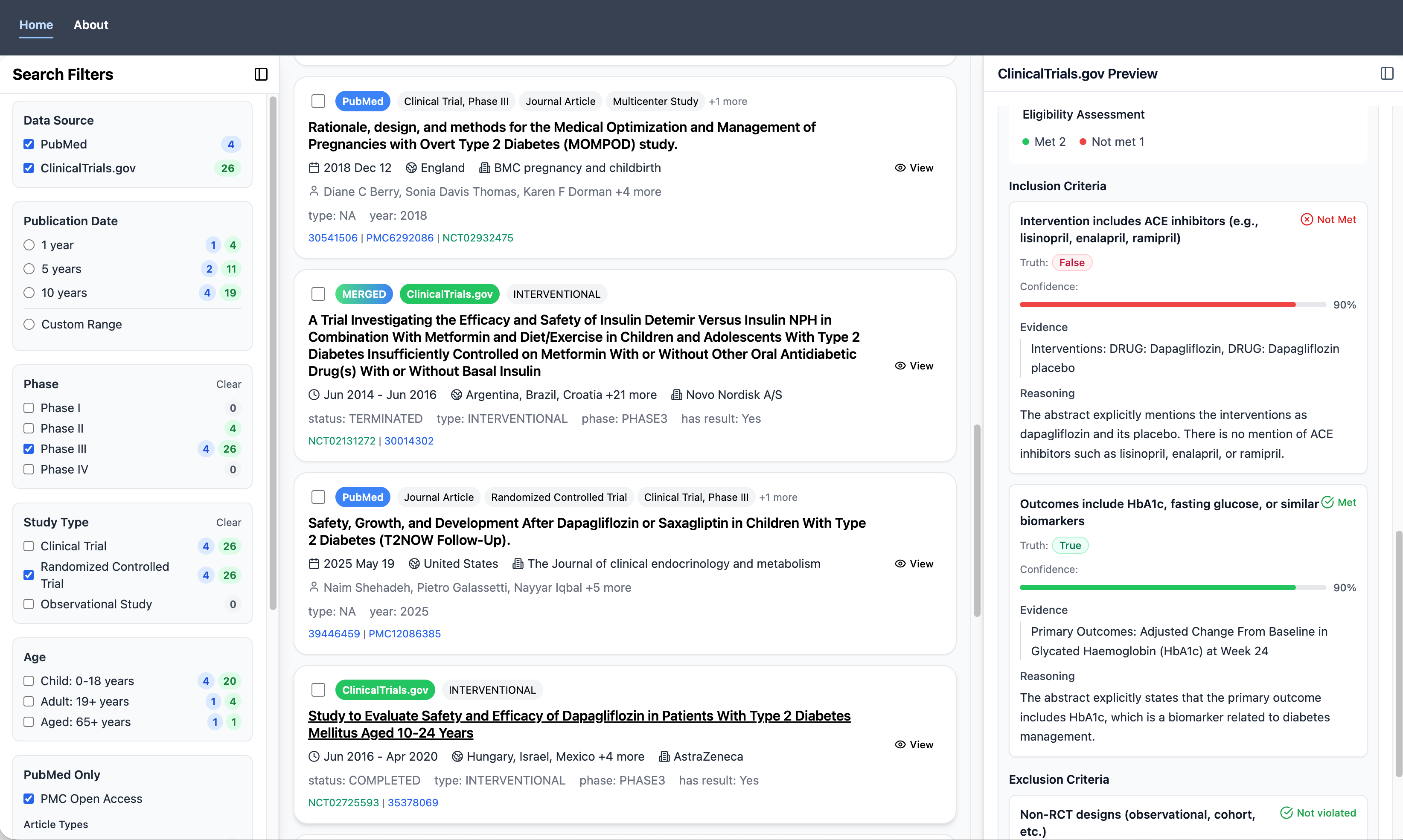}
  \caption{Eligibility check results}
  \label{fig:UI-SearchPage-EligibilityCriteria}
\end{figure}
\FloatBarrier
\begin{figure}[H]
  \centering
  \includegraphics[width=\textwidth]{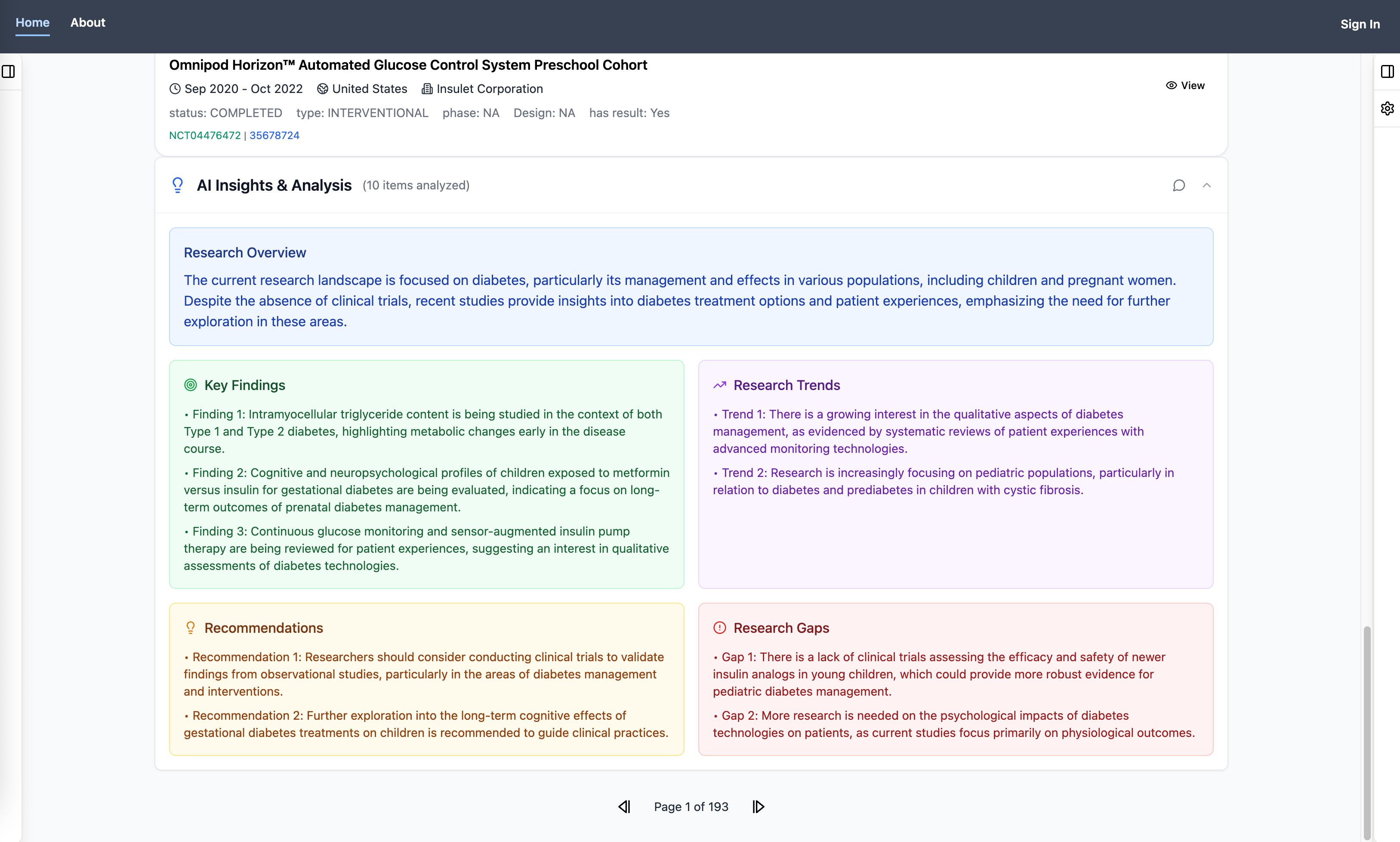}
  \caption{AI Insights}
  \label{fig:UI-SearchPage-AIInsight}
\end{figure}
\FloatBarrier
\section{User Interface - Detail Page}
\label{app:UI-DetailPage}
\begin{figure}[H]
  \centering
  \includegraphics[width=\textwidth]{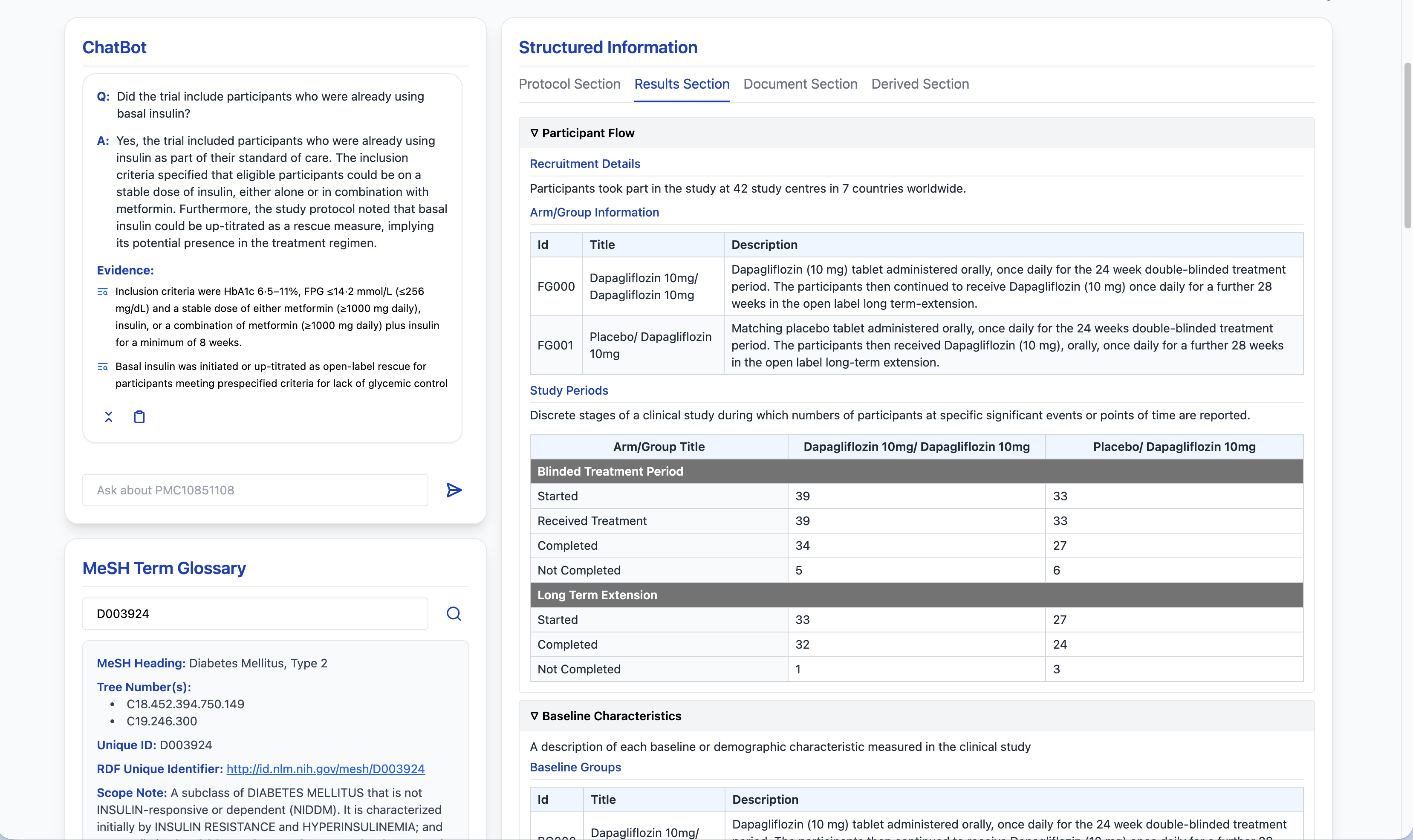}
  \caption{Detail page}
  \label{fig:UI-DetailPage}
\end{figure}
\begin{figure}[H]
  \centering
  \includegraphics[width=\textwidth]{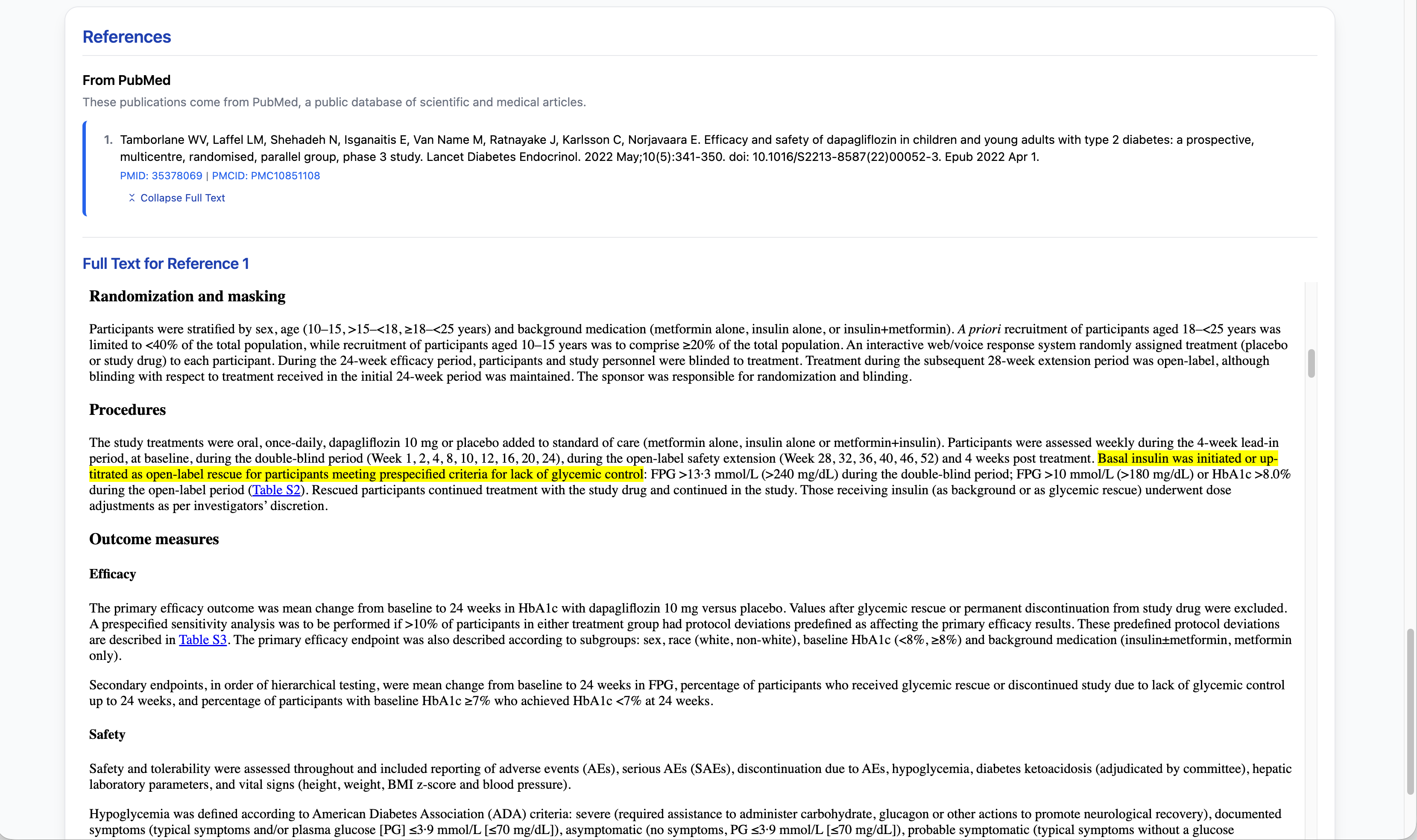}
  \caption{QA evidence highlight}
  \label{fig:UI-UI-DetailPage-ChatbotHighlight}
\end{figure}
\FloatBarrier
\clearpage

\section{Search Query Generation Prompt Templates}
\label{app:QueryGeneration-Prompt}
\begin{figure}[h]
  \centering
  \includegraphics[width=\textwidth]{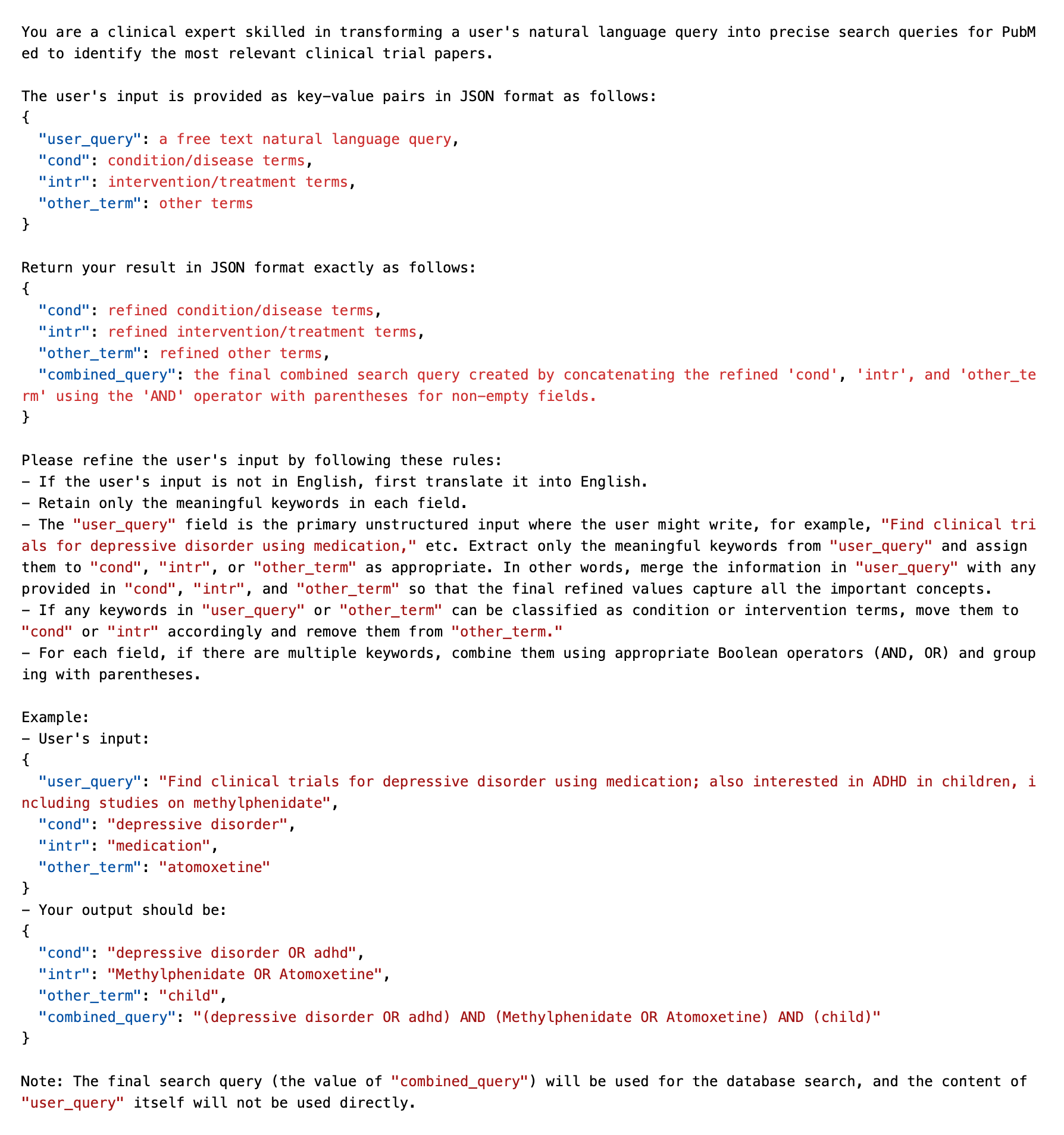}
  \caption{Search Query Generation - System Prompt}
  \label{fig:QueryGeneration-Prompt-System}
\end{figure}
\FloatBarrier
\clearpage

\begin{figure}[h]
  \centering
  \includegraphics[width=\textwidth]{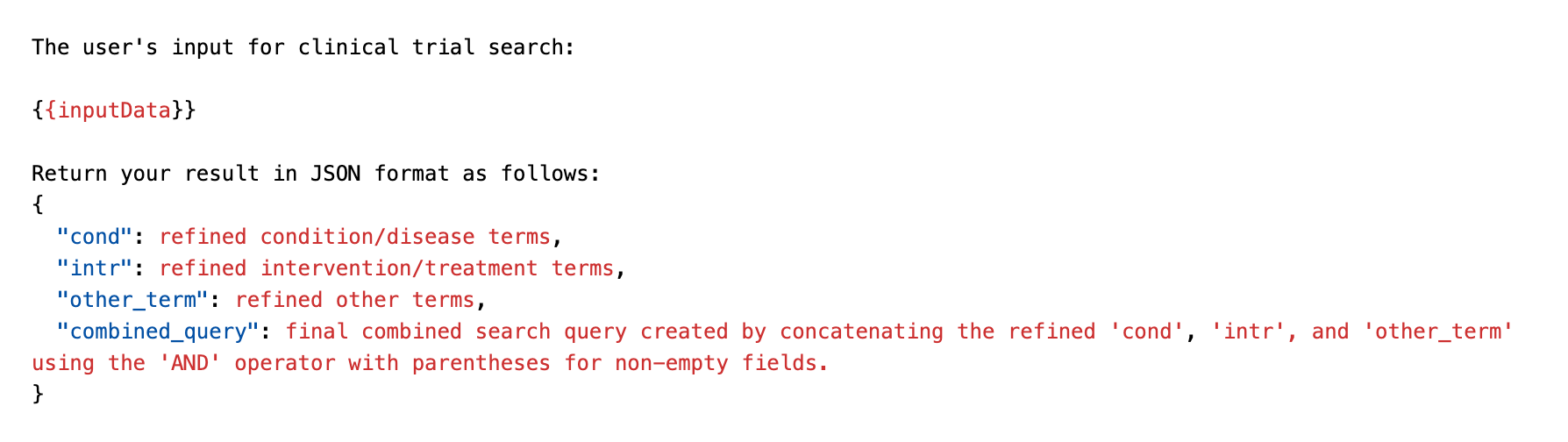}
  \caption{Search Query Generation - User Prompt}
  \label{fig:QueryGeneration-Prompt-User}
\end{figure}
\FloatBarrier

\section{Information Extraction Prompt Templates}
\label{app:IE-Prompt}
\begin{figure}[h]
  \centering
  \includegraphics[width=\textwidth]{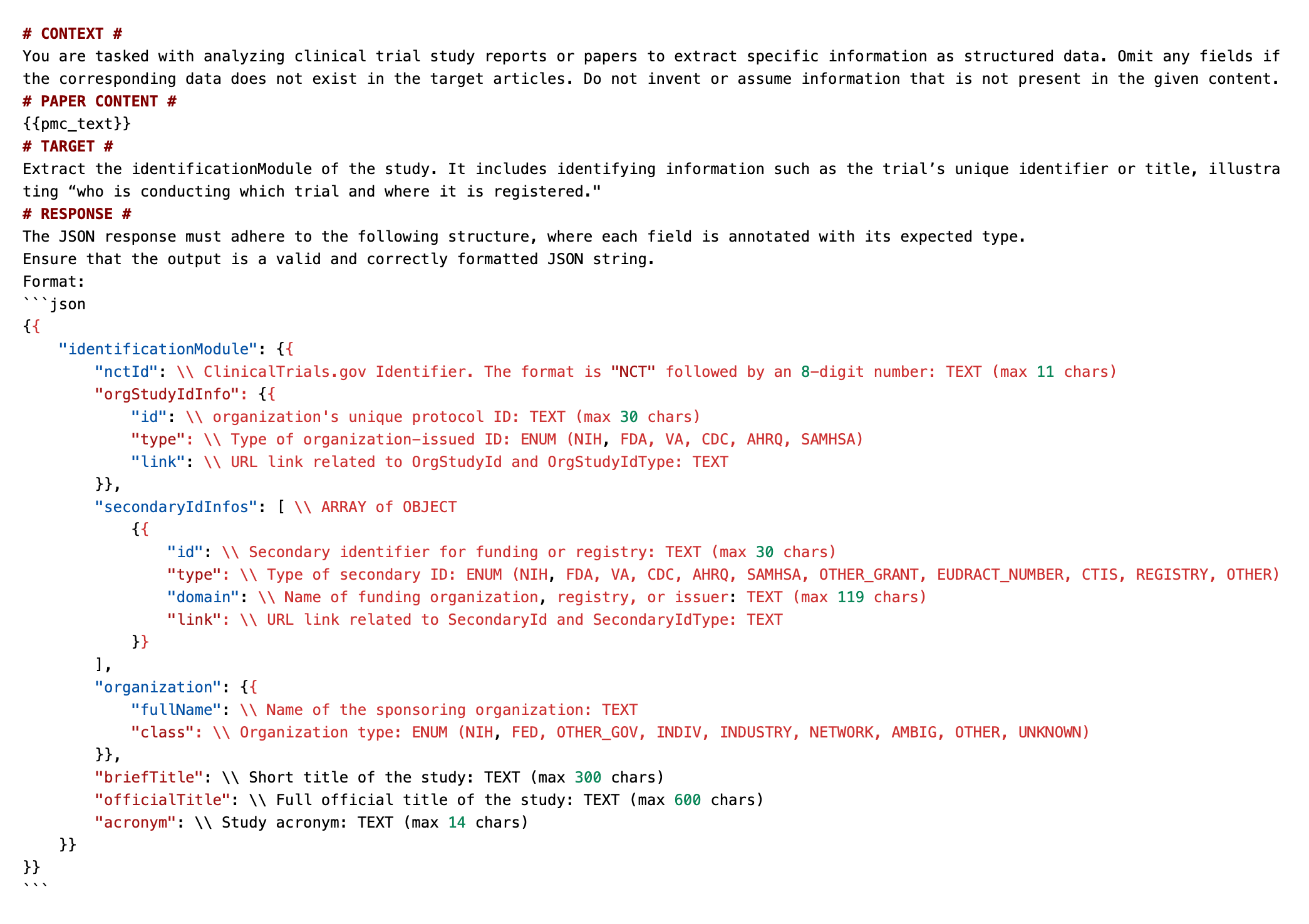}
  \caption{Protocol Section - Identification Module}
  \label{fig:IE-Prompt-1_Protocol_Identification}
\end{figure}
\begin{figure*}[h]
  \centering
  \includegraphics[width=\textwidth]{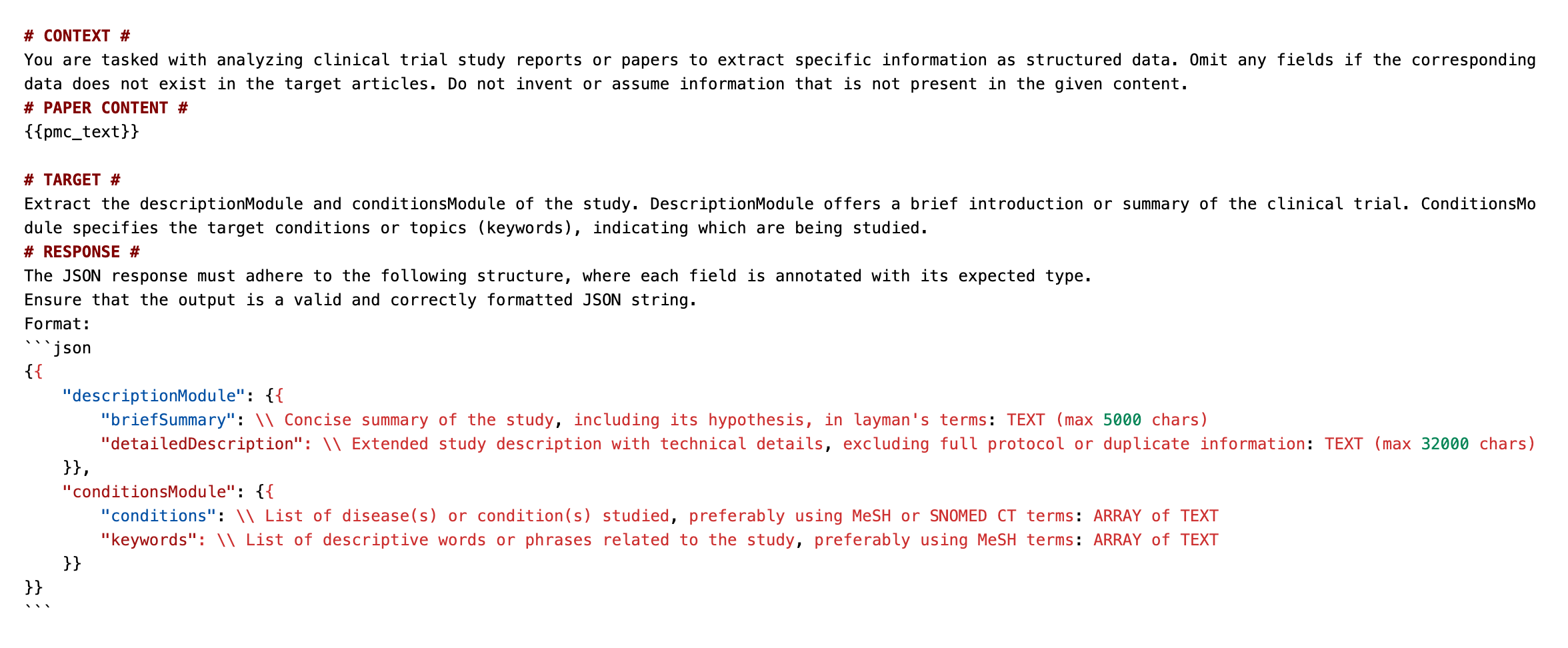}
  \caption{Protocol Section - Description Module, Conditions Module}
  \label{fig:IE-Prompt-2_Protocol_DescriptionAndConditions}
\end{figure*}
\begin{figure*}[h]
  \centering
  \includegraphics[width=\textwidth]{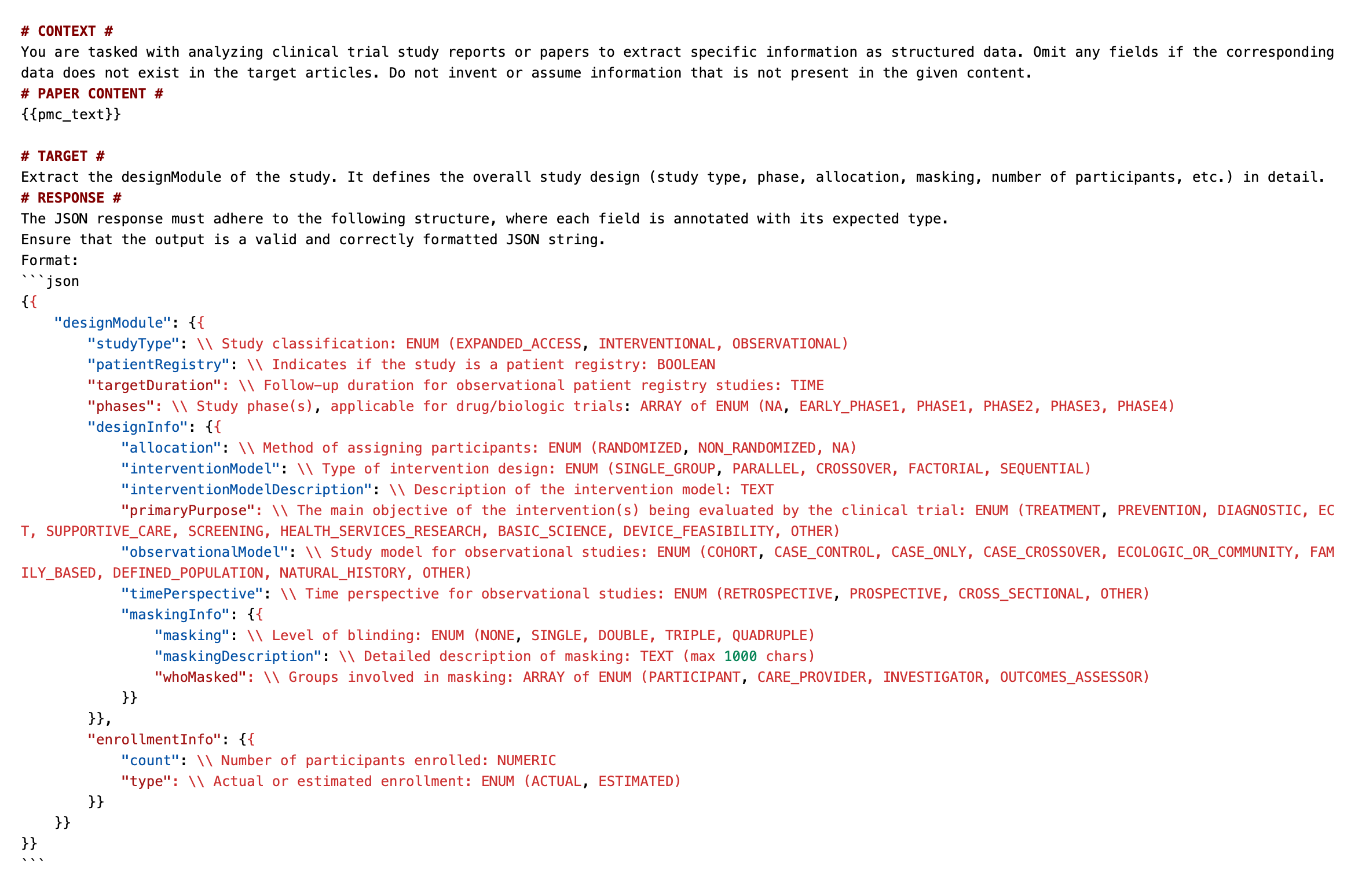}
  \caption{Protocol Section - Design Module}
  \label{fig:IE-Prompt-3_Protocol_Design}
\end{figure*}
\begin{figure*}[h]
  \centering
  \includegraphics[width=\textwidth]{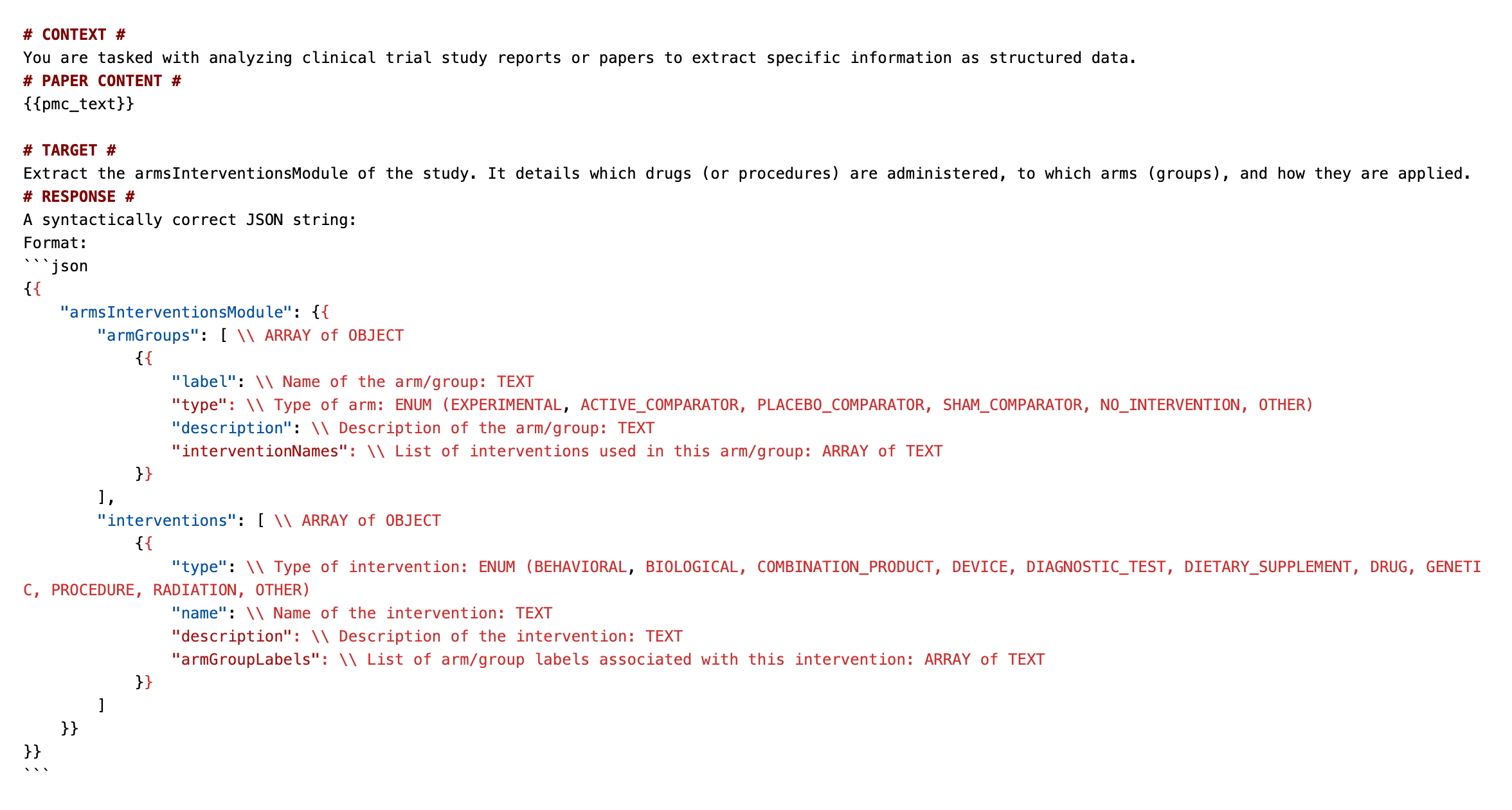}
  \caption{Protocol Section - Arms Interventions Module}
  \label{fig:IE-Prompt-4_Protocol_ArmsInterventions}
\end{figure*}
\begin{figure*}[h]
  \centering
  \includegraphics[width=\textwidth]{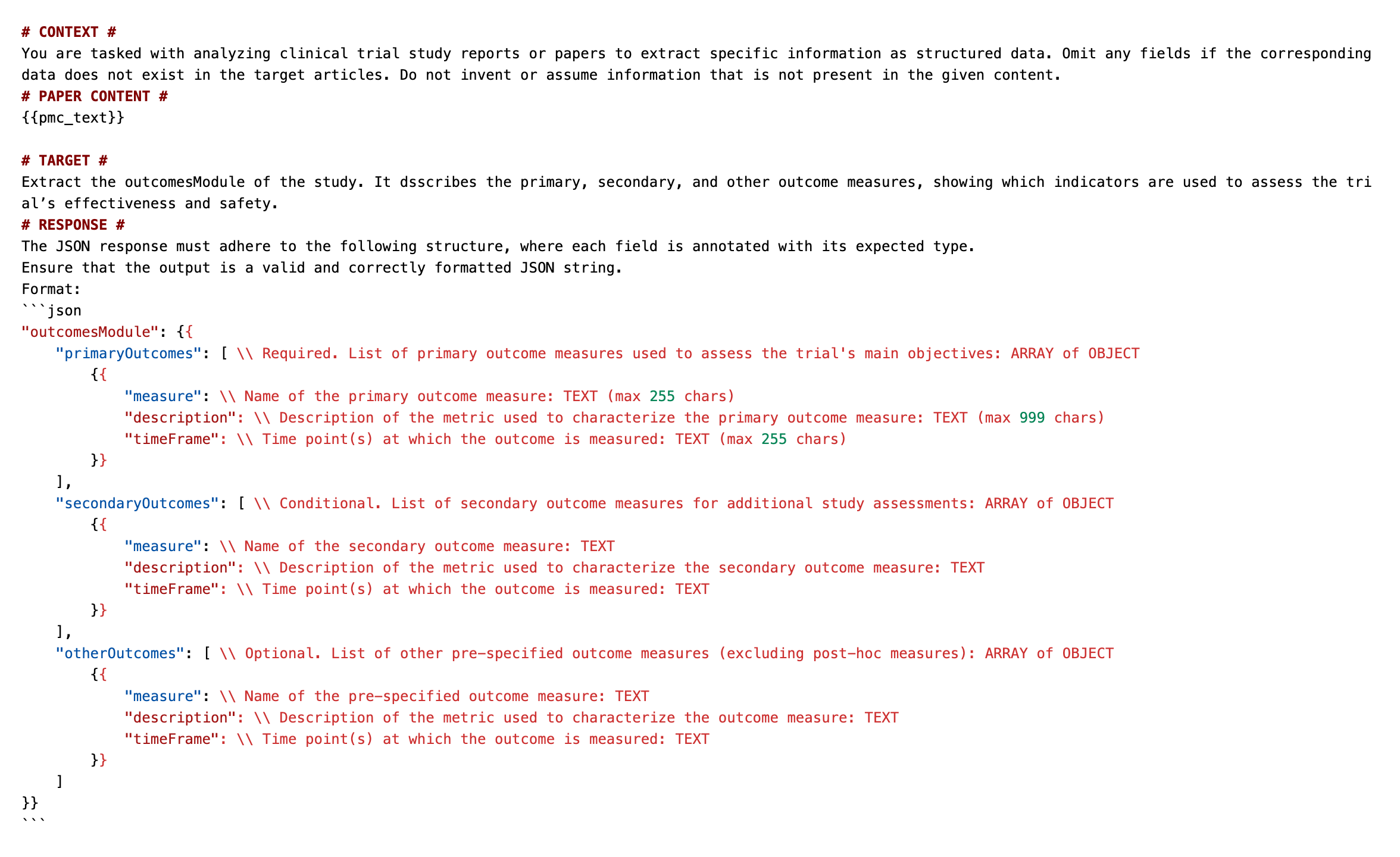}
  \caption{Protocol Section - Outcomes Module}
  \label{fig:IE-Prompt-5_Protocol_Outcomes}
\end{figure*}
\begin{figure*}[h]
  \centering
  \includegraphics[width=\textwidth]{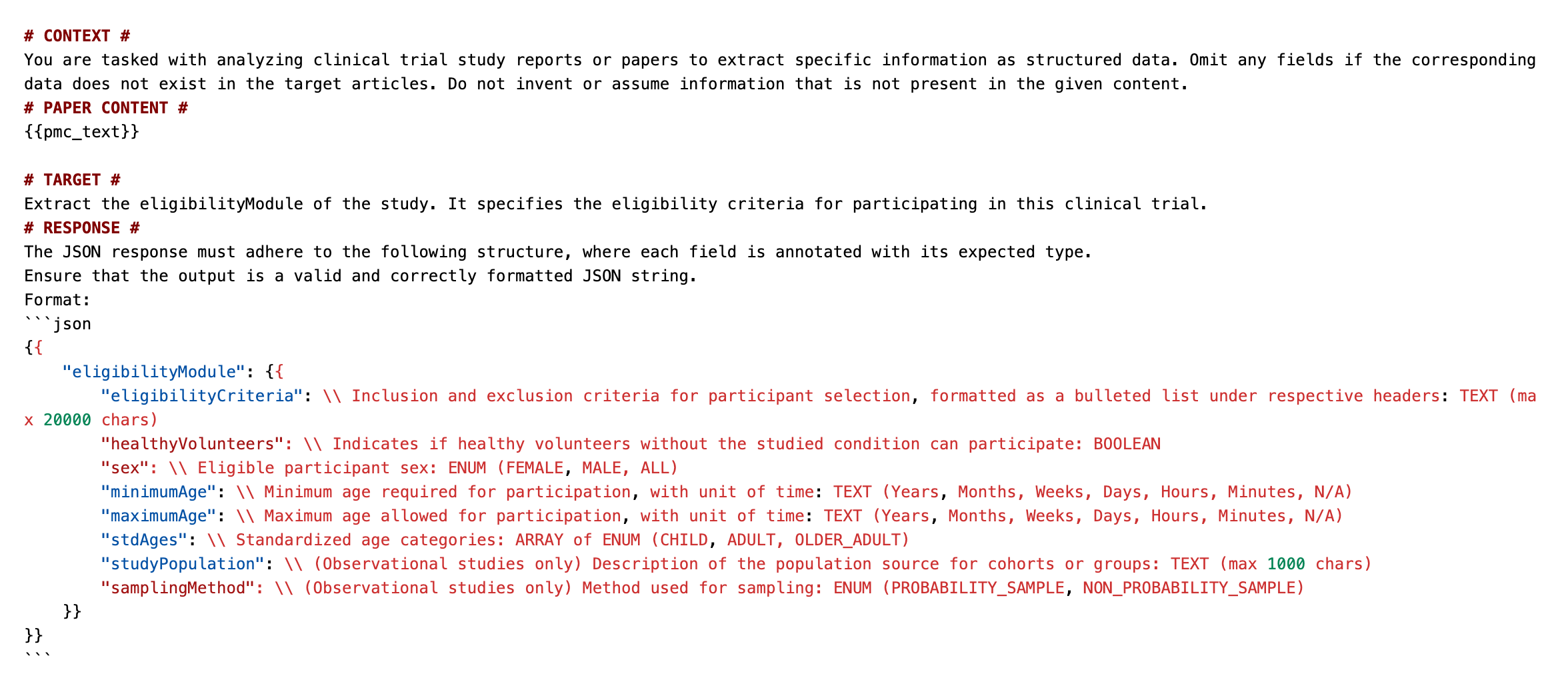}
  \caption{Protocol Section - Eligibility Module}
  \label{fig:IE-Prompt-6_Protocol_Eligibility}
\end{figure*}
\begin{figure*}[h]
  \centering
  \includegraphics[width=\textwidth]{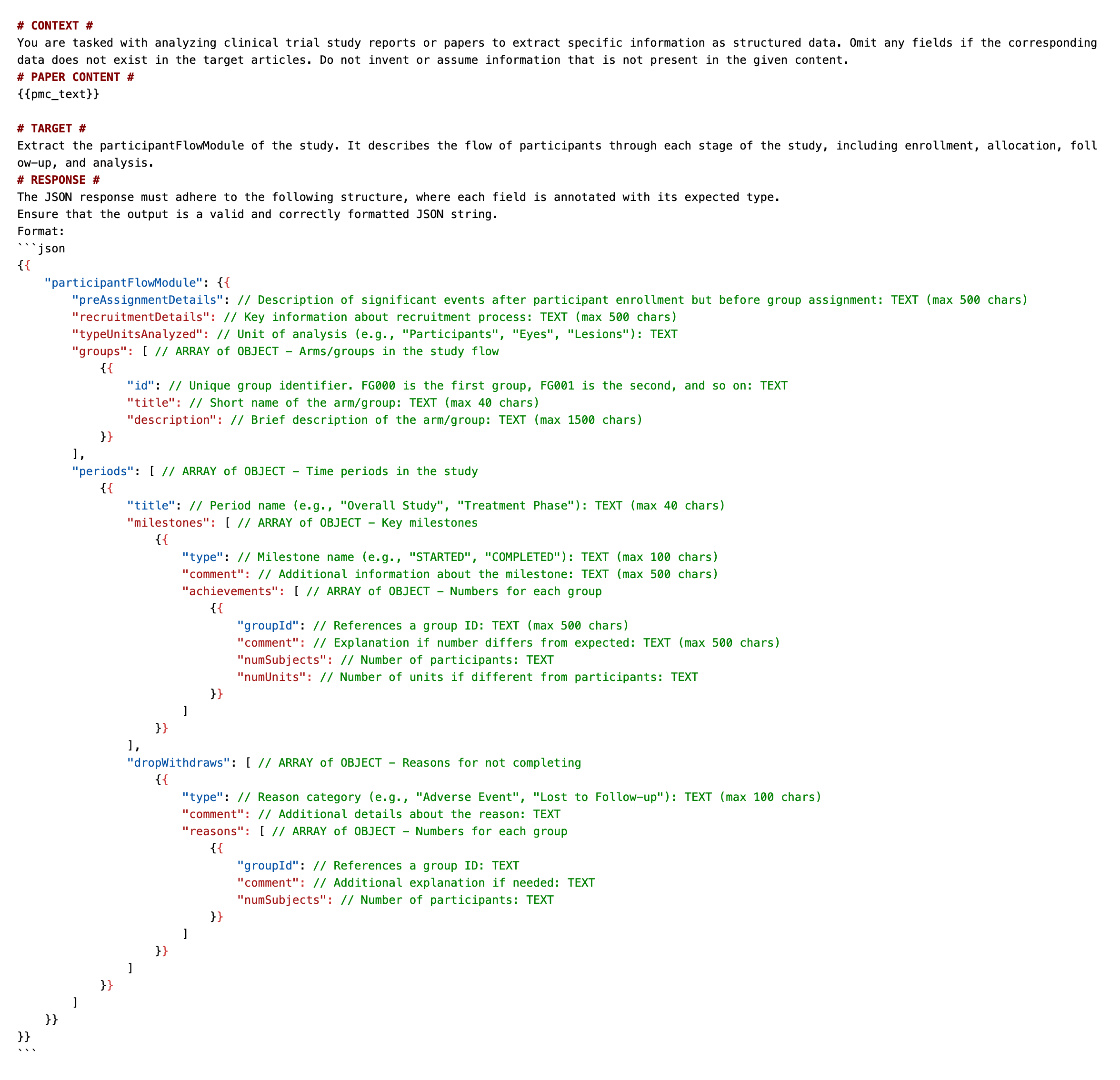}
  \caption{Results Section - Participant Flow Module}
  \label{fig:IE-Prompt-3_Protocol_Design}
\end{figure*}
\begin{figure*}[t]
  \centering
  \includegraphics[width=0.98\textwidth]{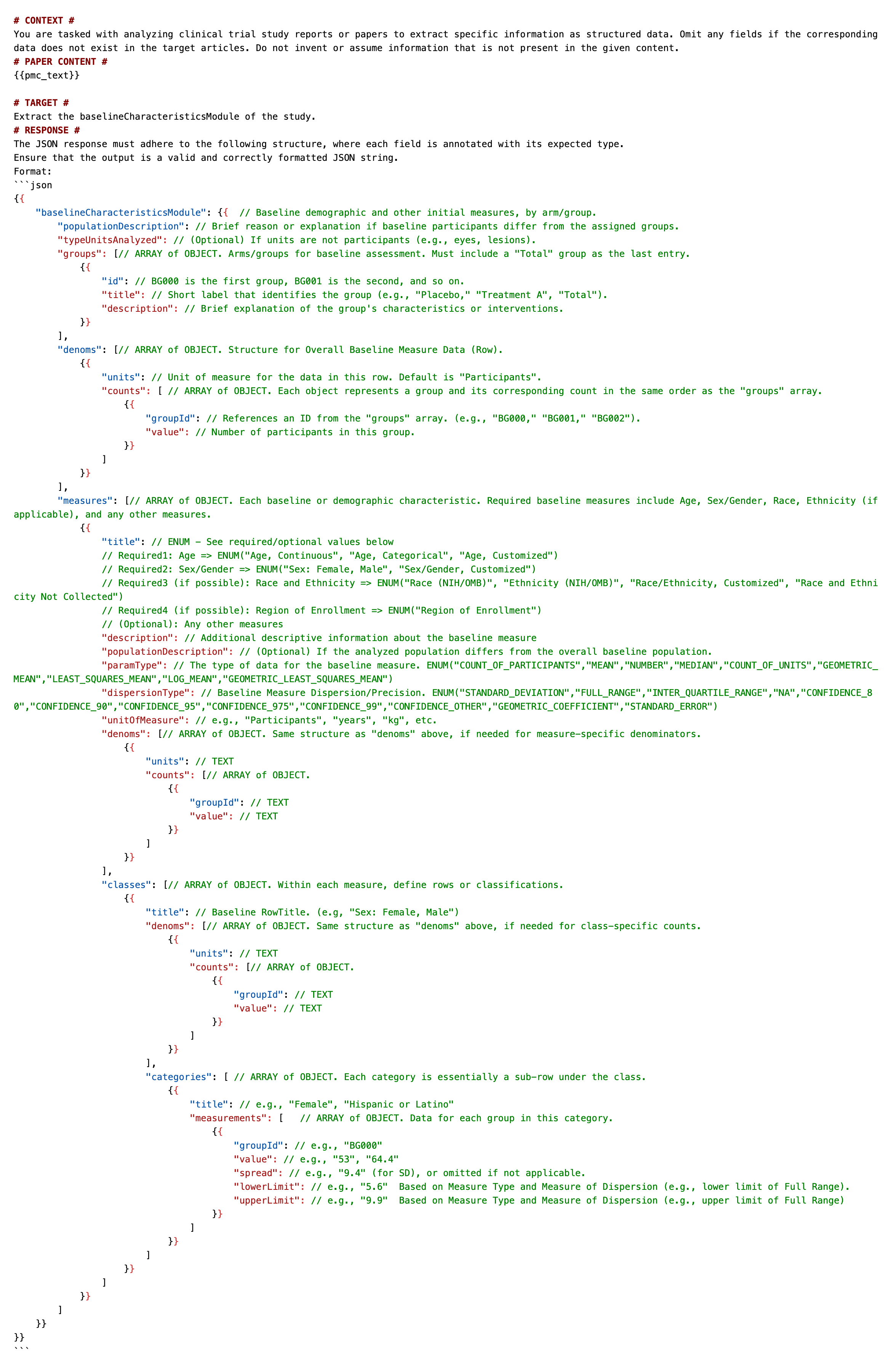}
  \caption{Results Section - Baseline Characteristics Module}
  \label{fig:IE-Prompt-8_Results_BaselineCharacteristics}
\end{figure*}
\begin{figure*}[h]
  \centering
  \includegraphics[width=\textwidth]{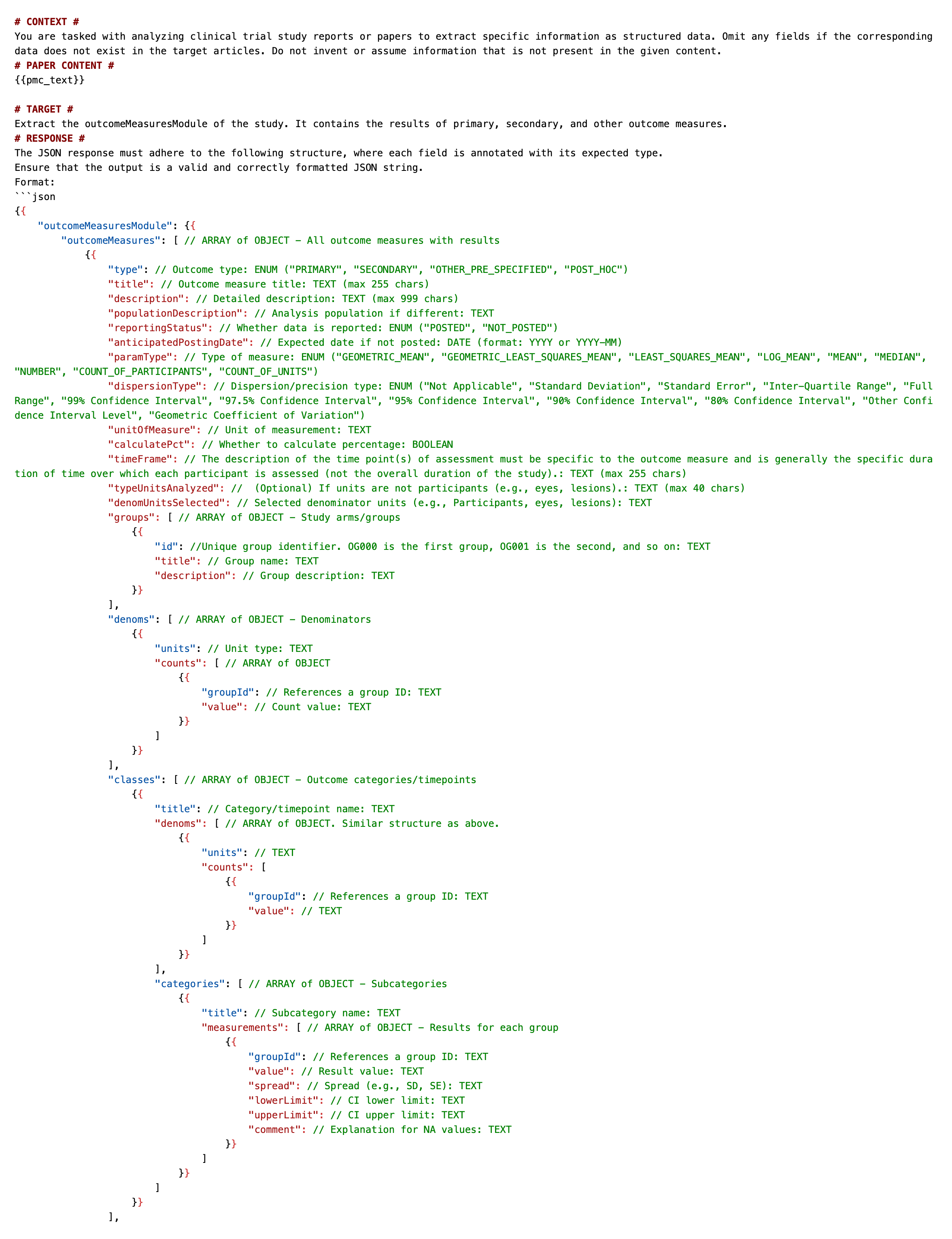}
  \caption{Results Section - Outcomes Measures Module 1}
  \label{IE-Prompt-9_Results_OutcomeMeasures1}
\end{figure*}
\begin{figure*}[h]
  \centering
  \includegraphics[width=\textwidth]{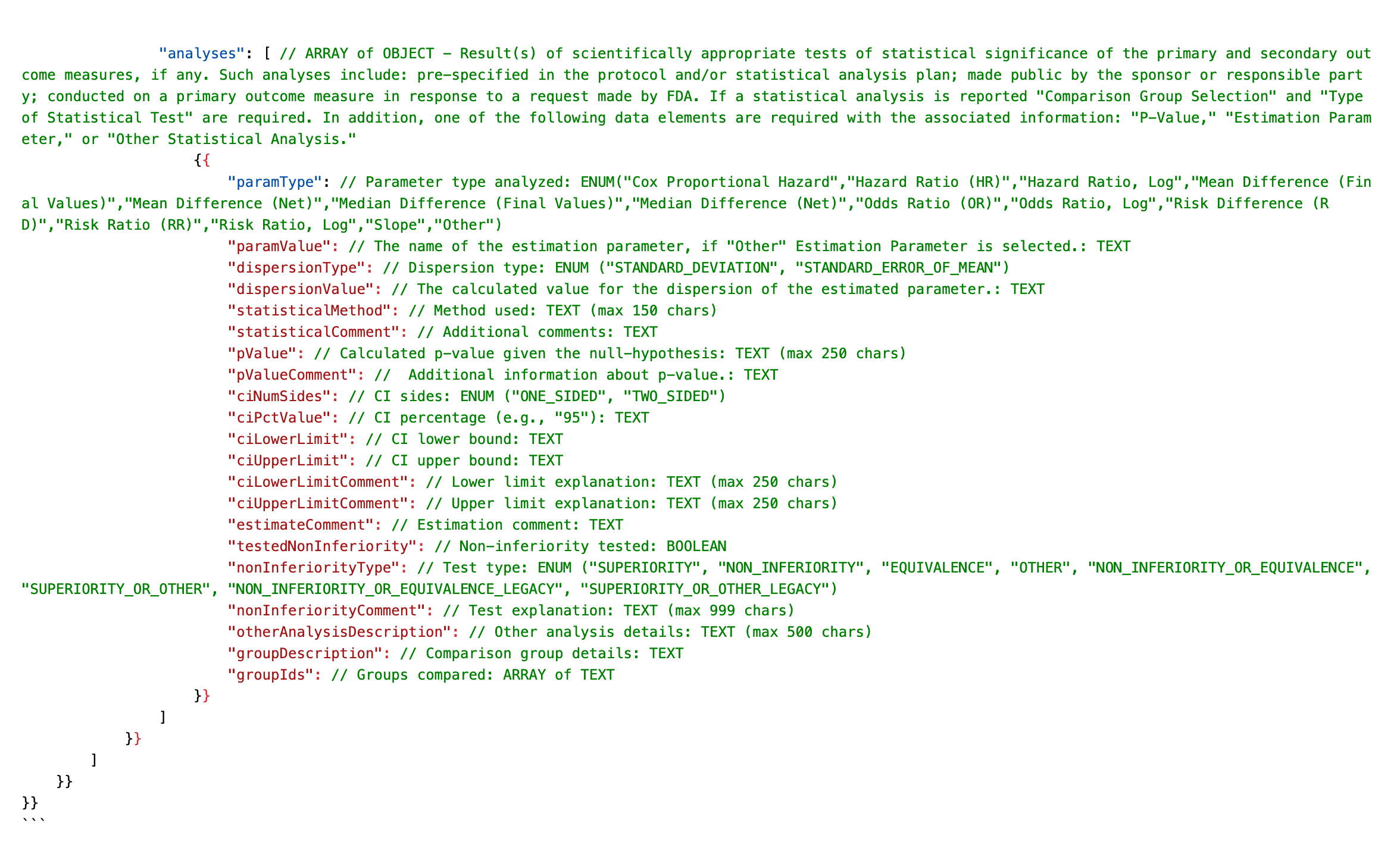}
  \caption{Results Section - Outcomes Measures Module 2}
  \label{fig:IE-Prompt-9_Results_OutcomeMeasures2}
\end{figure*}
\begin{figure*}[h]
  \centering
  \includegraphics[width=\textwidth]{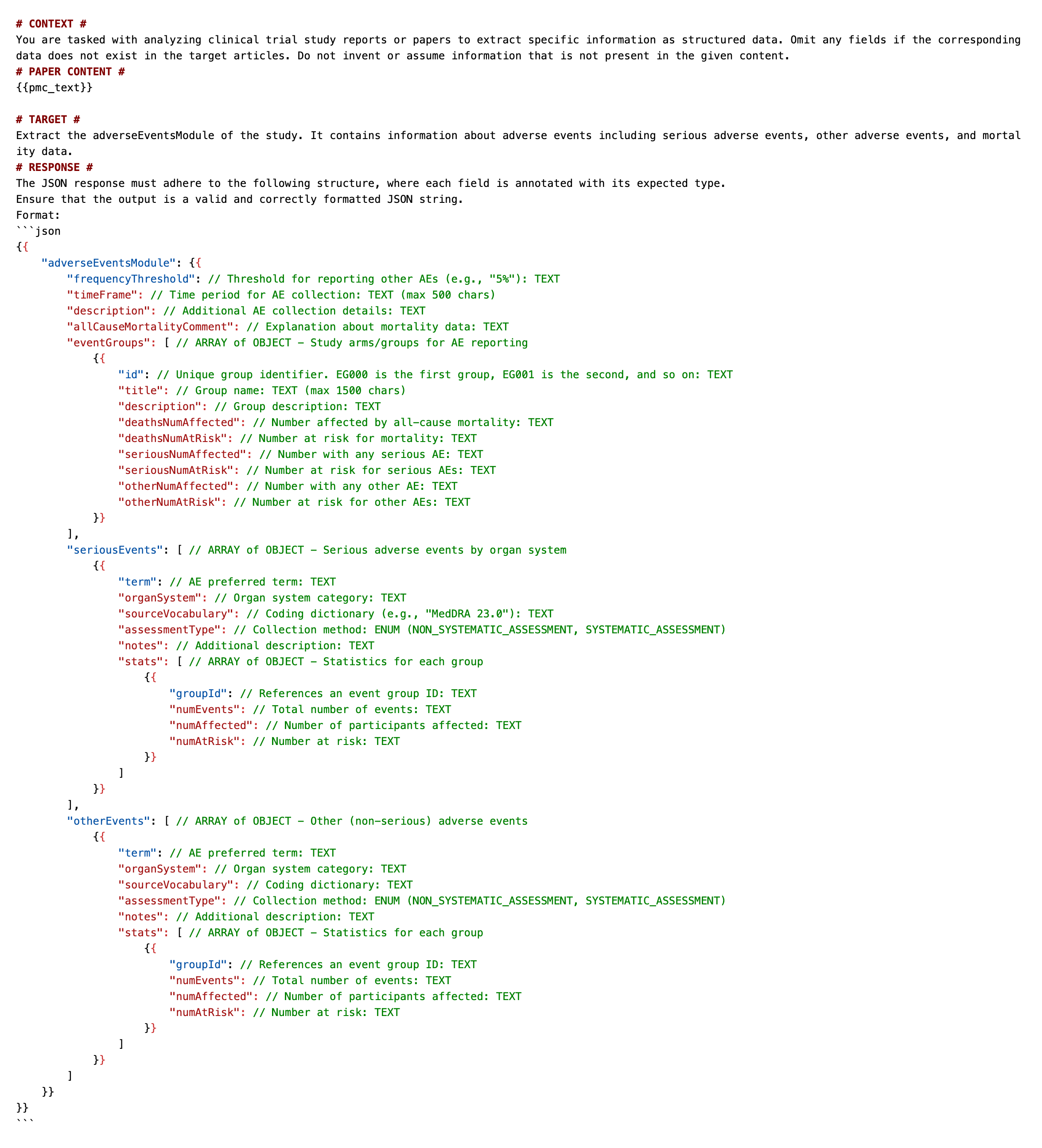}
  \caption{Results Section - Adverse Events Module}
  \label{fig:IE-Prompt-10_Results_AdverseEvents}
\end{figure*}
\begin{figure*}[h]
  \centering
  \includegraphics[width=\textwidth]{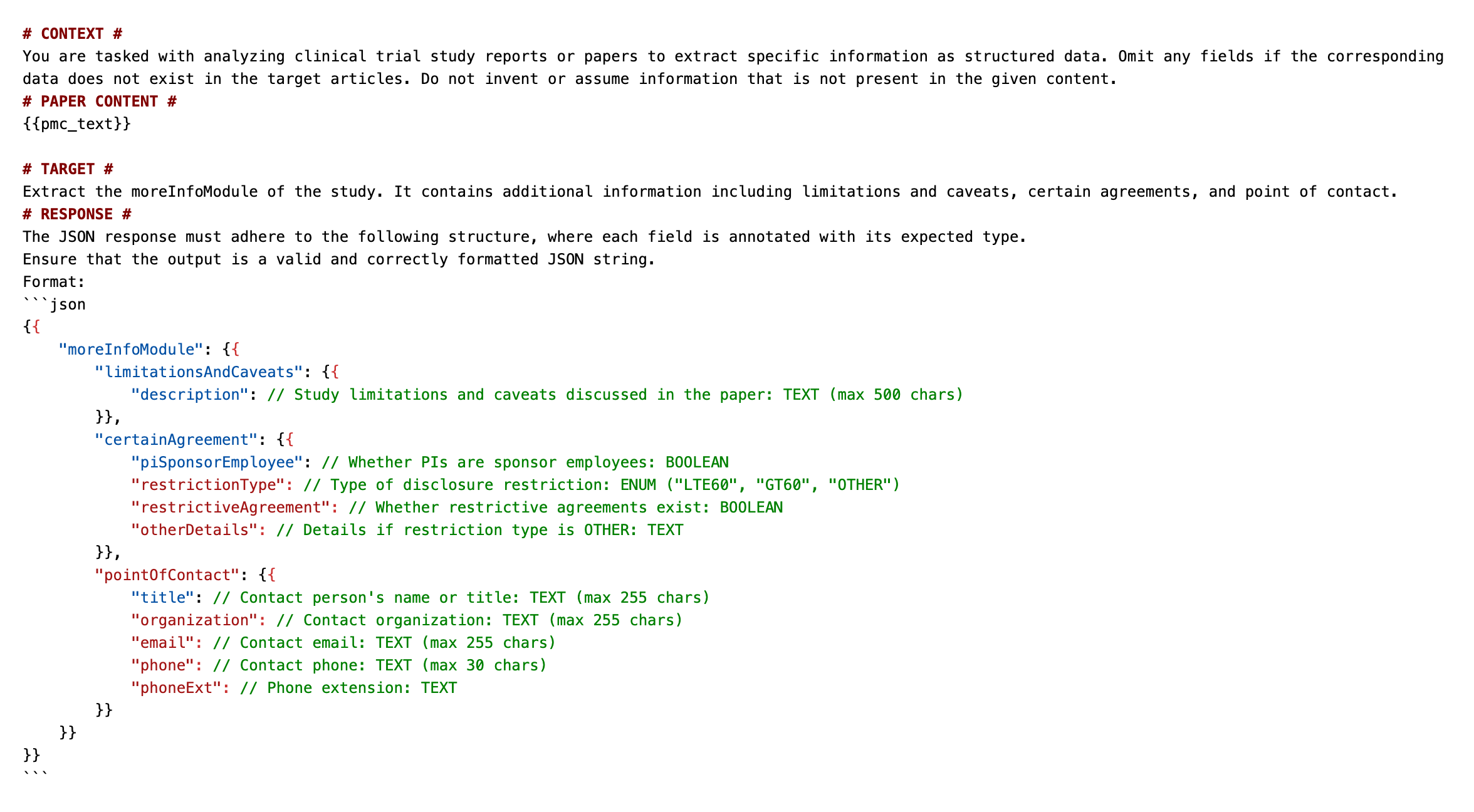}
  \caption{Results Section - More Info Module}
  \label{fig:IE-Prompt-11_Results_MoreInfo}
\end{figure*}
\begin{figure*}[h]
  \centering
  \includegraphics[width=\textwidth]{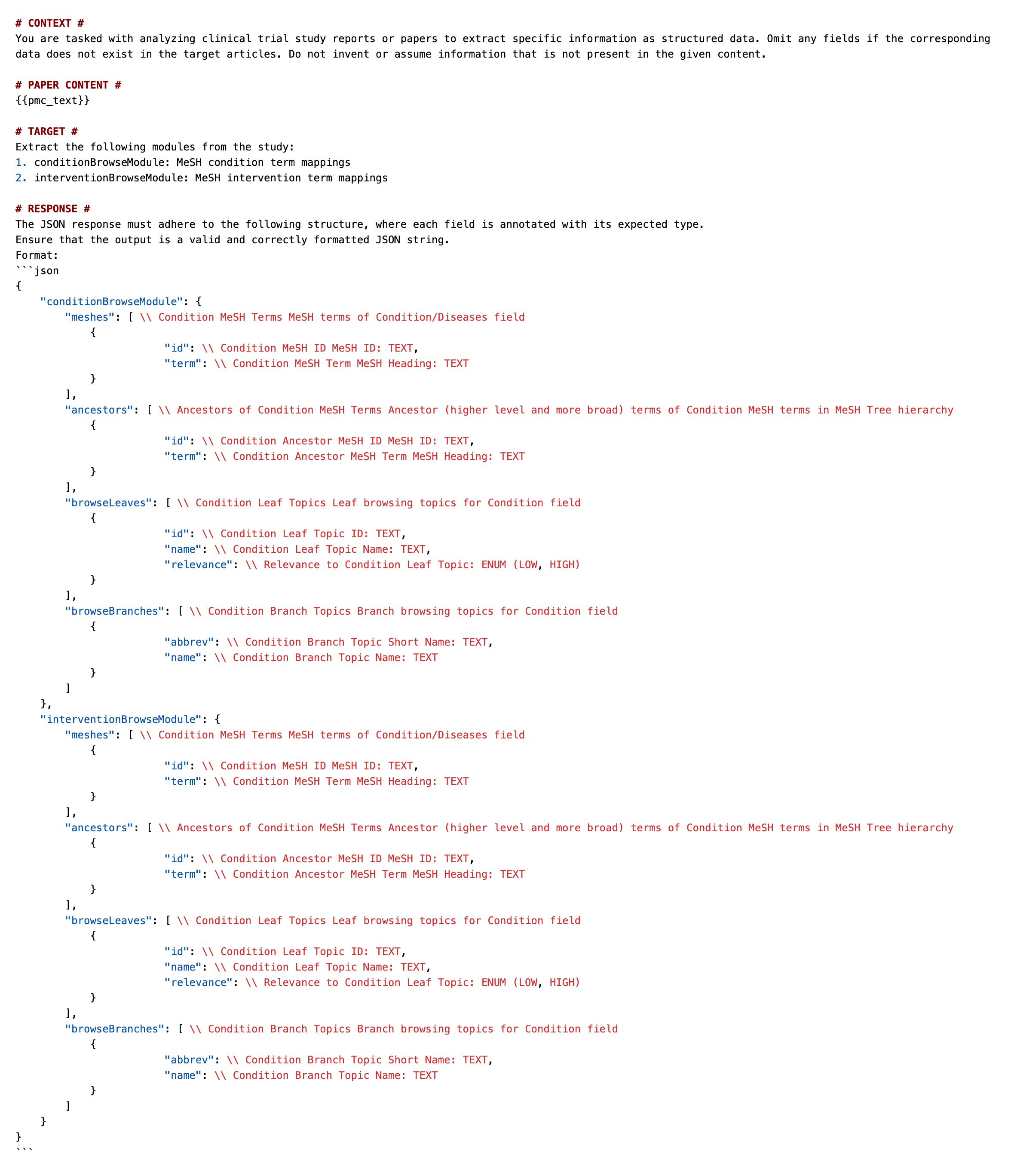}
  \caption{Results Section - Condition Browse Module, Intervention Browse Module}
  \label{fig:IE-Prompt-12_Results_ConditionBrowseInterventionBrowse}
\end{figure*}
\end{document}

%% file: tables/US-Scenarios.tex
\renewcommand{\arraystretch}{1.6}
\begin{table*}[t]
\centering
\small
\caption{Participant occupations and task contexts}
\label{tab:US-Scenarios}
\resizebox{\textwidth}{!}{
\begin{tabular}{@{}c p{4cm} p{11cm}@{}}
\toprule
\textbf{ID} & \textbf{Occupation} & \textbf{Task for This Evaluation} \\
\midrule
1 & MD, Specializing in Hematology &
I am the page editor for the Myelofibrosis evidence page on a hematology platform that aggregates and curates high-quality data from recent hematologic trials. My responsibility is to regularly review newly published clinical trials and update the page with evidence that meets my inclusion criteria. For this evaluation, I need to identify and organize the conclusions of recent myelofibrosis trials that should be added to the Myelofibrosis evidence page. \\
\hline
2 & Pharmacist &
I am looking for dosing decision evidence for a specific patient who is receiving cefepime and continuous renal replacement therapy (CRRT) in the setting of acute kidney failure, to ensure that the cefepime dose remains below the toxic range while still being effective. \\
\hline
3 & PhD Student -- pharmaceutical science &
I am preparing my candidacy proposal and reviewing preliminary studies exploring aspirin for the prevention of pre-eclampsia in high-risk pregnancies. I would like to examine how prior studies have evaluated placental biomarkers in this context. \\
\hline
4 & Statistician &
I am designing a clinical trial for a new EGFR-targeted therapy in non-small cell lung cancer. For this evaluation, I need to review existing evidence on EGFR-targeted therapies, including their efficacy and safety profiles, to support the study design. \\
\hline
5 & Nurse (PhD Candidate) &
I am conducting a systematic review of dyadic and family-based interventions for adults with type 2 diabetes, focusing on diabetes self-management, self-efficacy, and related psychosocial outcomes. My responsibility is to identify recent randomized and registrational trials, screen them against predefined inclusion criteria, and organize the conclusions with respect to dyadic or family-based diabetes self-management interventions. \\
\hline
6 & Dentist; General Dentistry &
I am preparing a paper that discusses the need to standardize IV moderate sedation training and competency assessment in dental residency programs. For this evaluation, I want to identify clinical trials and observational studies on IV moderate sedation in dentistry, including evidence on patient selection, resident training, and the appropriate order and quantity of medications for successful and safe IV moderate sedation. \\
\hline
7 & Pathologist (Dermatopathology / Gynecologic Pathology) &
I recently reviewed a case involving a metastatic cutaneous melanoma with a BRAFV600E mutation, confirmed through our molecular service. The patient has completed wide excision and sentinel lymph node evaluation, and the oncology team is now considering adjuvant systemic therapy options. For this case, I would like to review the clinical evidence supporting commonly used adjuvant therapies for metastatic melanoma so I can provide informed context when discussing the pathology findings with the treating oncologists. In particular, I want to find studies evaluating adjuvant systemic treatments used after complete resection of metastatic melanoma, including outcomes such as recurrence risk and treatment-related toxicity. \\
\bottomrule
\end{tabular}}
\end{table*}

%% file: tables/US-Search-SearchQueryGeneration.tex
{
\renewcommand{\arraystretch}{1.4}
\begin{table*}[t]
\centering
\small
\caption{Participant Natural Language Queries and Generated Structured Queries}
\label{tab:US-Search-SearchQueryGeneration}
\resizebox{\textwidth}{!}{
\begin{tabular}{c p{6cm} p{8cm}}
\toprule
\textbf{ID} & \textbf{Natural Language Query} & \textbf{Structured Query (CTH Output)} \\
\midrule

2 &
Randomized controlled trials involving continuous renal replacement therapy and cefepime in patients with acute kidney injury &
Condition: acute kidney injury \newline Intervention: continuous renal replacement therapy OR cefepime \newline Other terms: randomized controlled trials \\
\hline

7 &
I want clinical trials evaluating chemotherapy in high-stage melanoma, particularly those reporting BRAFV600E mutation status. &
Condition: high-stage melanoma OR BRAFV600E mutation \newline Intervention: chemotherapy \\

\bottomrule
\end{tabular}}
\end{table*}
}

%% file: tables/US-Search-Queries.tex
\renewcommand{\arraystretch}{1.4}
\begin{table*}[t]
\centering
\small
\caption{Participant Search Queries Across Systems}
\label{tab:US-Search-Queries}
\resizebox{\textwidth}{!}{
\begin{tabular}{c p{3.5cm} p{5cm} p{5cm}}
\toprule
\textbf{ID} & \textbf{PubMed Query} & \textbf{ClinicalTrials.gov Query} & \textbf{CTH Query} \\
\midrule

1 &
myelofibrosis OR polycythemia vera OR essential thrombocythemia OR myeloproliferative &
Condition: myelofibrosis OR polycythemia vera OR essential thrombocythemia OR myeloproliferative &
Condition: myelofibrosis OR polycythemia vera OR essential thrombocythemia OR myeloproliferative \\
\hline

2 &
cefepime and crrt &
Condition: Acute Kidney Injury (AKI)\newline
Intervention: Continuous Renal Replacement Therapy (CRRT)\newline
Other terms: cefepime &
--
\\
\hline

3 &
Aspirin AND Pre-eclampsia AND biomarkers &
Condition: Pre-eclampsia\newline
Intervention: aspirin\newline
Other terms: biomarkers &
Common: pregnancy\newline
Condition: pre-eclampsia\newline
Intervention: aspirin\newline
Other terms: biomarkers,\newline placental biomarkers \\
\hline

4 &
EGFR-targeted therapies non small cell lung cancer &
Condition: non small cell lung cancer\newline
Intervention: EGFR-targeted therapies\newline &
Common: “EGFR-targeted therapies non small cell lung cancer”\newline
Intervention: EGFR-targeted therapies \\
\hline

5 &
“type 2 diabetes” OR “type 2 diabetes mellitus” OR T2DM AND (dyadic OR family-based OR spouse OR partner OR dyadic OR caregiver OR self-management OR education) &
Condition: Type 2 Diabetes Mellitus (T2DM)\newline
Other terms: dyadic / family-based interventions, self-management outcomes &
Condition: type 2 diabetes mellitus\newline
Other terms: dyadic, family-based, spouse, partner, caregiver, self-management, education \\
\hline

6 &
IV moderate sedation AND midazolam AND fentanyl &
Other terms: IV moderate sedation AND midazolam AND fentanyl &
Other terms: Dental Anxiety AND Dental procedures AND IV moderate sedation guidelines \\
\hline

7 &
metastatic melanoma and BRAFV600E mutation and chemotherapy &
Condition: metastatic melanoma\newline
Intervention: chemotherapy\newline
Other terms: BRAF V600E mutation positive &
Common: cutaneous melanoma OR adjuvant melanoma OR resected melanoma OR "stage III melanoma" OR "BRAF V600E melanoma" \\

\bottomrule
\end{tabular}}
\end{table*}

%% file: tables/US-Search-Filters.tex
\renewcommand{\arraystretch}{1.4}
\begin{table*}[t]
\centering
\small
\caption{Participant Filters Used Across Systems}
\label{tab:US-Search-Filters}
\resizebox{\textwidth}{!}{
\begin{tabular}{c p{4.5cm} p{5.5cm} p{5cm}}
\toprule
\textbf{ID} & \textbf{PubMed Filters} & \textbf{CTG Filters} & \textbf{CTH Filters} \\
\midrule

1 &
Publication Date (Last 5 years)\newline
Phase (III/IV)\newline
Article Type (Clinical Trial, RCT)\newline
Access (PMC OA)\newline
Species (Humans)
&
Completion Date (Last 5 years)\newline
Phase (III/IV)\newline
Study Type (Interventional)\newline
Has Results (True)\newline
Status (Completed)
&
Date (Last 5 years)\newline
Phase (III/IV)\newline
Study Type (Clinical Trial, RCT)\newline
PubMed (PMC OA, Humans)\newline
CTG (With Results, Completed)
\\
\hline

2 &
Age\newline
Publication Date
&
Age\newline
Completion Date
&
Age\newline
Date
\\
\hline

3 &
Publication Date
&
Has Results (True)
&
Phase\newline
Study Type
\\
\hline

4 &
Publication Date (Last 10 years)\newline
Article Type (Clinical Trial)
&
Completion Date (From:11/17/2015)
&
Date (Last 10 years)\newline
Study Type (Clinical Trial)
\\
\hline

5 &
Publication Date (Last 10 years)\newline
Article Type (Clinical Trial, RCT)\newline
Species (Humans)\newline
Age (Adults)\newline
Language (English)
&
Completion Date (Last 10 years)\newline
Study Type (Interventional, Randomized)\newline
Status (Completed)\newline
Has Results (True)\newline
Age (Adults)
&
Date (Last 10 years)\newline
Study Type (Clinical Trial, RCT)\newline
Age (Adults)\newline
PubMed (PMC OA, Humans)\newline
CTG (With Results, Completed)
\\

\bottomrule
\end{tabular}}
\end{table*}

%% file: tables/US-Search-EligibiligyCheck.tex
\renewcommand{\arraystretch}{1.4}
\begin{table*}[t]
\centering
\small
\caption{Participant Inclusion and Exclusion Criteria}
\label{tab:US-Search-EligibiligyCheck}
\resizebox{\textwidth}{!}{
\begin{tabular}{c p{7cm} p{7cm}}
\toprule
\textbf{ID} & \textbf{Inclusion Criteria} & \textbf{Exclusion Criteria} \\
\midrule

1 &
Randomized controlled or single-arm registrational trials with $\geq$50 patients \newline
\noindent\textcolor{gray}{\rule{\linewidth}{0.1pt}}
Reported at least one primary outcome
&
Other myeloproliferative neoplasms without fibrosis (e.g., PV or ET without MF) \newline
\noindent\textcolor{gray}{\rule{\linewidth}{0.1pt}}
Prefibrotic MF or early myeloproliferative disease without confirmed fibrosis
\\
\hline

2 &
AKI \newline
\noindent\textcolor{gray}{\rule{\linewidth}{0.1pt}}
CRRT
&
\\
\hline

5 &
Randomized controlled trials or single-arm registrational trials with approximately $\geq$50 participants \newline
\noindent\textcolor{gray}{\rule{\linewidth}{0.1pt}}
Adults ($\geq$18 years) with type 2 diabetes mellitus \newline
\noindent\textcolor{gray}{\rule{\linewidth}{0.1pt}}
Dyadic or family-based behavioral, psychoeducational, or self-management intervention involving partner/spouse/family \newline
\noindent\textcolor{gray}{\rule{\linewidth}{0.1pt}}
Reported at least one primary or key outcome (self-efficacy, self-management behaviors, dyadic/family processes, HbA1c, etc.) \newline
\noindent\textcolor{gray}{\rule{\linewidth}{0.1pt}}
Follow-up $\geq$3 months after baseline
&
Studies focusing only on type 1 diabetes, gestational diabetes, or prediabetes \newline
\noindent\textcolor{gray}{\rule{\linewidth}{0.1pt}}
Purely pharmacologic comparisons without behavioral/educational/family components \newline
\noindent\textcolor{gray}{\rule{\linewidth}{0.1pt}}
Pediatric/adolescent-only populations \newline
\noindent\textcolor{gray}{\rule{\linewidth}{0.1pt}}
Case reports, very small pilot studies, qualitative-only studies without quantitative outcomes
\\
\hline

6 &
Dental
&
General Anesthesia
\\

\bottomrule
\end{tabular}}
\end{table*}

%% file: tables/US-ChatbotInteractions.tex
\renewcommand{\arraystretch}{1.25}
\begin{table*}[t]
\centering
\small
\caption{Participant Chatbot Interactions}
\label{tab:US-Chatbot}
\resizebox{\textwidth}{!}{
\begin{tabular}{@{}c p{15cm}@{}}
\toprule
\textbf{ID} & \textbf{Chatbot Interaction} \\
\midrule

1 &
\textbf{PMC12092244 | What was the exact pelabresib dose and schedule?}\newline
Pelabresib was administered orally at a starting dose of 125\,mg once daily for 14 consecutive days followed by a 7-day break, forming a 21-day cycle. Dose modifications between 50--175\,mg once daily were permitted based on adverse events and treatment response.\newline
\noindent\textcolor{gray}{\rule{\linewidth}{0.3pt}}
\textbf{PMC12092244 | What was the primary endpoint, and was it met?}\newline
The primary endpoint was achieving a \(\geq 35\%\) reduction in spleen volume from baseline at week~24. This endpoint was met, with a significantly greater proportion of patients in the pelabresib--ruxolitinib arm achieving this reduction compared to the placebo--ruxolitinib arm.\newline
\noindent\textcolor{gray}{\rule{\linewidth}{0.3pt}}
\textbf{PMC12092244 | What were the most common adverse events?}\newline
The most common treatment-emergent adverse events were hematological—primarily thrombocytopenia and anemia. Among non-hematological events, diarrhea was most frequent in the pelabresib--ruxolitinib arm, whereas constipation was most frequent in the placebo--ruxolitinib arm.
\\

\midrule

2 &
\textbf{PMC10576861 | Since ACORN showed higher neurotoxicity with Cefepime, should we reconsider de-escalating from Piperacillin-Tazobactam to Cefepime solely based on renal concerns?}\newline
Based on the ACORN trial findings, reconsidering the practice of de-escalating from Piperacillin-Tazobactam to Cefepime solely for renal protection is warranted. The study demonstrated that there was no significant difference in the highest stage of acute kidney injury or death between the two treatment groups. Furthermore, the trial concluded that Piperacillin-Tazobactam did not increase the incidence of acute kidney injury, whereas Cefepime was associated with a higher incidence of neurological dysfunction.\newline
\noindent\textcolor{gray}{\rule{\linewidth}{0.3pt}}\newline
\textbf{PMC9463324 | The study showed creatinine elevation but no difference in dialysis or mortality. Does this mean the current KDIGO criteria are too sensitive for antibiotic-induced AKI?}\newline
The study suggests that the creatinine-based KDIGO criteria may identify "pseudotoxicity" rather than true renal injury in the context of Vancomycin and Piperacillin-Tazobactam therapy. The authors propose that the observed creatinine elevation is likely due to the inhibition of tubular secretion rather than damage to the kidney parenchyma, a conclusion supported by the lack of corresponding changes in alternative biomarkers (cystatin C and BUN) and the absence of increased risk for dialysis or mortality.
\\
\midrule

3 &
\textbf{NCT03893630 | Did the study report any differences in placental biomarker levels between the 81mg and 162mg aspirin groups?}\newline
Yes, the study reported specific mean changes in placental biomarker levels for the 81mg and 162mg aspirin groups, although statistical significance of the differences was not explicitly detailed in the provided text. For Placental Growth Factor (PIGF), the mean change from baseline to Event 2 was 568.3 pg/ml for the 81mg group and 509.7 pg/ml for the 162mg group. For soluble Intercellular Adhesion Molecule (s-ICAM), the mean change from baseline to Event 2 was 1.6 ng/ml for the 81mg group and 7.9 ng/ml for the 162mg group. Data for IL-6 and TNF were not reported due to insufficient serum samples.
\\
\midrule

4 &
\textbf{PMC12139608 | Can you summarize how the sample size is determined in this study?}\newline
The study powered for $\geq$ 85\% to detect a 25\% difference in ORR using one-sided $\alpha$=0.025; planned enrollment was ~126 patients.\newline
\noindent\textcolor{gray}{\rule{\linewidth}{0.3pt}}
\textbf{PMC10318754 | Can you summarize the statistical analysis methods used?}\newline
Analyses included chi-square/Fisher tests and QoL correlations with PFS/OS and sponsorship.\newline
\noindent\textcolor{gray}{\rule{\linewidth}{0.3pt}}
\textbf{PMC9509093 | Summarize tabulated results for all endpoints.}\newline
Aumolertinib improved PFS (19.3 vs 9.9 mo); ORR 73.8\% vs 72.1\%; DCR 93.0\% vs 96.7\%; DoR 18.1 vs 8.3 mo; fewer rash/diarrhea events.
\\
\midrule

5 &
\textbf{PMC8266295 | Were any clinical outcomes such as HbA1c reported, and if so, what were the main results?}\newline
The paper does not report clinical outcomes such as HbA1c changes as part of the results. The study focused on family supporter involvement, support techniques, and caregiving experience rather than direct clinical outcomes like HbA1c.
\\
\midrule

6 &
\textbf{PMC5858011 | What do they recommend as the safest order?}\newline
Midazolam-first caused fewer desaturation events (90–92\%).
\\
\midrule

7 &
\textbf{NCT01682083 | If OS wasn’t clearly significant early on, is the decision to use dabrafenib/trametinib driven mostly by the strong RFS benefit?}\newline
Yes, the provided results indicate that the primary outcome, Relapse-free Survival (RFS), demonstrated a statistically significant benefit with a hazard ratio of 0.47 and a p-value of < 0.0001. In contrast, the Overall Survival (OS) analysis resulted in a p-value of 0.006, which did not meet the pre-specified significance threshold of 0.000019 for the first interim analysis. \newline
\\
\bottomrule
\end{tabular}}
\end{table*}

%% file: tables/ie-eval-fields.tex
\begin{table}[H]
\centering
\footnotesize
\setlength{\tabcolsep}{4pt}
\begin{tabular}{lcc|ccc}
\toprule
\textbf{Piece} & \textbf{\# Ref} & \textbf{\# EV} &
\textbf{Gemini} & \textbf{GPT} & \textbf{Claude} \\ 
\midrule
DesignObservationalModel & 1 & 7.67 & \textbf{1.00} & \textbf{1.00} & \textbf{1.00} \\
\textbf{DetailedDescription} & \textbf{64} & \textbf{31.67} &
\textbf{1.00} & \textbf{1.00} & 0.86 \\
StudyType & 100 & 0.00 & \textbf{1.00} & \textbf{1.00} & 0.85 \\
DesignAllocation & 99 & 0.67 & 0.99 & \textbf{1.00} & 0.85 \\
BriefSummary & 100 & 0.00 & \textbf{1.00} & \textbf{1.00} & 0.83 \\
Condition & 100 & 0.00 & \textbf{1.00} & \textbf{1.00} & 0.83 \\
DesignInterventionModel & 99 & 0.33 & 0.98 & \textbf{1.00} & 0.85 \\
Phase & 99 & 0.67 & 0.98 & \textbf{1.00} & 0.85 \\
EnrollmentCount & 100 & 0.00 & 0.99 & 0.99 & 0.84 \\
DesignMasking & 99 & 0.67 & 0.98 & 0.99 & 0.85 \\
NCTId & 100 & 0.00 & 0.99 & \textbf{1.00} & 0.82 \\
\textbf{Keyword} & \textbf{52} & \textbf{44.00} &
\textbf{1.00} & \textbf{1.00} & 0.80 \\
MinimumAge & 99 & 0.67 & 0.99 & 0.99 & 0.77 \\
EligibilityCriteria & 100 & 0.00 & 0.99 & 0.99 & 0.77 \\
HealthyVolunteers & 99 & 1.00 & 0.99 & 0.99 & 0.76 \\
Sex & 100 & 0.00 & 0.99 & 0.99 & 0.77 \\
\textbf{MaximumAge} & \textbf{54} & \textbf{39.00} &
0.99 & \textbf{1.00} & 0.74 \\
StdAge & 100 & 0.00 & 0.98 & 0.95 & 0.76 \\
DesignWhoMasked & 59 & 3.67 & 0.88 & \textbf{0.94} & 0.77 \\
BriefTitle & 100 & 0.00 & 0.71 & \textbf{1.00} & 0.84 \\
OfficialTitle & 100 & 0.00 & 0.71 & 0.81 & 0.82 \\
\bottomrule
\end{tabular}
\captionof{table}{
Average Extra-Valid (EV) case analysis and information extraction performance for each field. 
\textbf{\# Ref}: number of evaluation cases (out of 100) where the field appears in CTG ground-truth. 
\textbf{\# EV}: average cases across models where the field was Extra-Valid (present in PMC but missing from CTG). 
Values $\ge 10$ are bolded. 
Highest F1 per row is bolded. Rows ordered by average F1.
}
\label{tab:ie-eval-fields}
\end{table}

%% file: tables/ie-demo-fields.tex
\footnotesize
\begin{longtable}{@{}%
  >{\raggedleft\arraybackslash}p{0.7cm}
  p{5.0cm}
  p{9.8cm}@{}}

\caption{Comprehensive list of all fields included in our information-extraction schema.}
\label{tab:ie-demo-fields}\\

\toprule
Index & Piece & Field Index \\
\midrule
\endfirsthead

\multicolumn{3}{c}%
{\tablename\ \thetable\ -- \textit{Continued from previous page}} \\
\toprule
Index & Piece & Field Index \\
\midrule
\endhead

\midrule
\multicolumn{3}{r}{\small Continued on next page} \\
\endfoot

\bottomrule
\endlastfoot

\multicolumn{3}{@{}l}{\textbf{protocolSection.identificationModule}}\\[1ex]
1     & NCTId                                   & nctId                                                       \\
2     & OrgStudyId                              & orgStudyIdInfo.id                                           \\
3     & OrgStudyIdType                          & orgStudyIdInfo.type                                         \\
4     & OrgStudyIdLink                          & orgStudyIdInfo.link                                         \\
5     & SecondaryId                             & secondaryIdInfos.id                                         \\
6     & SecondaryIdType                         & secondaryIdInfos.type                                       \\
7     & SecondaryIdLink                         & secondaryIdInfos.link                                       \\
8     & BriefTitle                              & briefTitle                                                  \\
9     & OfficialTitle                           & officialTitle                                               \\
10    & Acronym                                 & acronym                                                     \\
11    & OrgFullName                             & organization.fullName                                       \\
12    & OrgClass                                & organization.class                                          \\
[1ex]\multicolumn{3}{@{}l}{\textbf{protocolSection.descriptionModule}}\\[1ex]
13    & BriefSummary                            & briefSummary                                                   \\
14    & DetailedDescription                     & detailedDescription                                            \\
[1ex]\multicolumn{3}{@{}l}{\textbf{protocolSection.conditionsModule}}\\[1ex]
15    & Condition                               & conditions                                                      \\
16    & Keyword                                 & keywords                                                        \\
[1ex]\multicolumn{3}{@{}l}{\textbf{protocolSection.designModule}}\\[1ex]
17    & StudyType                               & studyType                                                           \\
18    & PatientRegistry                         & patientRegistry                                                     \\
19    & TargetDuration                          & targetDuration                                                      \\
20    & Phase                                   & phases                                                              \\
21    & DesignAllocation                        & designInfo.allocation                                               \\
22    & DesignInterventionModel                 & designInfo.interventionModel                                        \\
23    & DesignInterventionModelDescription      & designInfo.interventionModelDescription                             \\
24    & DesignPrimaryPurpose                    & designInfo.primaryPurpose                                           \\
25    & DesignObservationalModel                & designInfo.observationalModel                                       \\
26    & DesignTimePerspective                   & designInfo.timePerspective                                          \\
27    & DesignMasking                           & designInfo.maskingInfo.masking                                      \\
28    & DesignMaskingDescription                & designInfo.maskingInfo.maskingDescription                           \\
29    & DesignWhoMasked                         & designInfo.maskingInfo.whoMasked                                    \\
30    & EnrollmentCount                         & enrollmentInfo.count                                                \\
31    & EnrollmentType                          & enrollmentInfo.type                                                 \\
[1ex]\multicolumn{3}{@{}l}{\textbf{protocolSection.armsInterventionsModule}}\\[1ex]
32    & ArmGroupLabel                           & armGroups.label                                          \\
33    & ArmGroupType                            & armGroups.type                                           \\
34    & ArmGroupDescription                     & armGroups.description                                    \\
35    & ArmGroupInterventionName                & armGroups.interventionNames                              \\
36    & InterventionType                        & interventions.type                                       \\
37    & InterventionName                        & interventions.name                                       \\
38    & InterventionDescription                 & interventions.description                                \\
39    & InterventionArmGroupLabel               & interventions.armGroupLabels                             \\
[1ex]\multicolumn{3}{@{}l}{\textbf{protocolSection.outcomesModule}}\\[1ex]
40    & PrimaryOutcomeMeasure                   & primaryOutcomes.measure                                           \\
41    & PrimaryOutcomeDescription               & primaryOutcomes.description                                       \\
42    & PrimaryOutcomeTimeFrame                 & primaryOutcomes.timeFrame                                         \\
43    & SecondaryOutcomeMeasure                 & secondaryOutcomes.measure                                         \\
44    & SecondaryOutcomeDescription             & secondaryOutcomes.description                                     \\
45    & SecondaryOutcomeTimeFrame               & secondaryOutcomes.timeFrame                                       \\
46    & OtherOutcomeMeasure                     & otherOutcomes.measure                                             \\
47    & OtherOutcomeDescription                 & otherOutcomes.description                                         \\
48    & OtherOutcomeTimeFrame                   & otherOutcomes.timeFrame                                           \\
[1ex]\multicolumn{3}{@{}l}{\textbf{protocolSection.eligibilityModule}}\\[1ex]
49    & EligibilityCriteria                     & eligibilityCriteria                                            \\
50    & HealthyVolunteers                       & healthyVolunteers                                              \\
51    & Sex                                     & sex                                                            \\
52    & MinimumAge                              & minimumAge                                                     \\
53    & MaximumAge                              & maximumAge                                                     \\
54    & StdAge                                  & stdAges                                                        \\
55    & StudyPopulation                         & studyPopulation                                                \\
56    & SamplingMethod                          & samplingMethod                                                 \\
[1ex]\multicolumn{3}{@{}l}{\textbf{resultsSection.participantFlowModule}}\\[1ex]
57    & FlowPreAssignmentDetails                & preAssignmentDetails                                        \\
58    & FlowRecruitmentDetails                  & recruitmentDetails                                          \\
59    & FlowTypeUnitsAnalyzed                   & typeUnitsAnalyzed                                           \\
60    & FlowGroupId                             & groups.id                                                   \\
61    & FlowGroupTitle                          & groups.title                                                \\
62    & FlowGroupDescription                    & groups.description                                          \\
63    & FlowPeriodTitle                         & periods.title                                               \\
64    & FlowMilestoneType                       & periods.milestones.type                                     \\
65    & FlowMilestoneComment                    & periods.milestones.comment                                  \\
66    & FlowAchievementGroupId                  & periods.milestones.achievements.groupId                     \\
67    & FlowAchievementComment                  & periods.milestones.achievements.comment                     \\
68    & FlowAchievementNumSubjects              & periods.milestones.achievements.numSubjects                 \\
69    & FlowAchievementNumUnits                 & periods.milestones.achievements.numUnits                    \\
70    & FlowDropWithdrawType                    & periods.dropWithdraws.type                                  \\
71    & FlowDropWithdrawComment                 & periods.dropWithdraws.comment                               \\
72    & FlowReasonGroupId                       & periods.dropWithdraws.reasons.groupId                       \\
73    & FlowReasonComment                       & periods.dropWithdraws.reasons.comment                       \\
74    & FlowReasonNumSubjects                   & periods.dropWithdraws.reasons.numSubjects                   \\
[1ex]\multicolumn{3}{@{}l}{\textbf{resultsSection.baselineCharacteristicsModule}}\\[1ex]
75    & BaselinePopulationDescription           & populationDescription                               \\
76    & BaselineTypeUnitsAnalyzed               & typeUnitsAnalyzed                                   \\
77    & BaselineGroupId                         & groups.id                                           \\
78    & BaselineGroupTitle                      & groups.title                                        \\
79    & BaselineGroupDescription                & groups.description                                  \\
80    & BaselineDenomUnits                      & denoms.units                                        \\
81    & BaselineDenomCountGroupId               & denoms.counts.groupId                               \\
82    & BaselineDenomCountValue                 & denoms.counts.value                                 \\
83    & BaselineMeasureTitle                    & measures.title                                      \\
84    & BaselineMeasureDescription              & measures.description                                \\
85    & BaselineMeasurePopulationDescription    & measures.populationDescription                      \\
86    & BaselineMeasureParamType                & measures.paramType                                  \\
87    & BaselineMeasureDispersionType           & measures.dispersionType                             \\
88    & BaselineMeasureUnitOfMeasure            & measures.unitOfMeasure                              \\
89    & BaselineMeasureDenomUnits               & measures.denoms.units                               \\
90    & BaselineMeasureDenomCountGroupId        & measures.denoms.counts.groupId                      \\
91    & BaselineMeasureDenomCountValue          & measures.denoms.counts.value                        \\
92    & BaselineClassTitle                      & measures.classes.title                              \\
93    & BaselineClassDenomUnits                 & measures.classes.denoms.units                       \\
94    & BaselineClassDenomCountGroupId          & measures.classes.denoms.counts.groupId              \\
95    & BaselineClassDenomCountValue            & measures.classes.denoms.counts.value                \\
96    & BaselineCategoryTitle                   & measures.classes.categories.title                   \\
97    & BaselineMeasurementGroupId              & measures.classes.categories.measurements.groupId    \\
98    & BaselineMeasurementValue                & measures.classes.categories.measurements.value      \\
99    & BaselineMeasurementSpread               & measures.classes.categories.measurements.spread     \\
100   & BaselineMeasurementLowerLimit           & measures.classes.categories.measurements.lowerLimit \\
101   & BaselineMeasurementUpperLimit           & measures.classes.categories.measurements.upperLimit \\
[1ex]\multicolumn{3}{@{}l}{\textbf{resultsSection.outcomeMeasuresModule}}\\[1ex]
102   & OutcomeMeasureType                      & outcomeMeasures.type                                        \\
103   & OutcomeMeasureTitle                     & outcomeMeasures.title                                       \\
104   & OutcomeMeasureDescription               & outcomeMeasures.description                                 \\
105   & OutcomeMeasurePopulationDescription     & outcomeMeasures.populationDescription                       \\
106   & OutcomeMeasureReportingStatus           & outcomeMeasures.reportingStatus                             \\
107   & OutcomeMeasureAnticipatedPostingDate    & outcomeMeasures.anticipatedPostingDate                      \\
108   & OutcomeMeasureParamType                 & outcomeMeasures.paramType                                   \\
109   & OutcomeMeasureDispersionType            & outcomeMeasures.dispersionType                              \\
110   & OutcomeMeasureUnitOfMeasure             & outcomeMeasures.unitOfMeasure                               \\
111   & OutcomeMeasureCalculatePct              & outcomeMeasures.calculatePct                                \\
112   & OutcomeMeasureTimeFrame                 & outcomeMeasures.timeFrame                                   \\
113   & OutcomeMeasureTypeUnitsAnalyzed         & outcomeMeasures.typeUnitsAnalyzed                           \\
114   & OutcomeMeasureDenomUnitsSelected        & outcomeMeasures.denomUnitsSelected                          \\
115   & OutcomeGroupId                          & outcomeMeasures.groups.id                                   \\
116   & OutcomeGroupTitle                       & outcomeMeasures.groups.title                                \\
117   & OutcomeGroupDescription                 & outcomeMeasures.groups.description                          \\
118   & OutcomeDenomUnits                       & outcomeMeasures.denoms.units                                \\
119   & OutcomeDenomCountGroupId                & outcomeMeasures.denoms.counts.groupId                       \\
120   & OutcomeDenomCountValue                  & outcomeMeasures.denoms.counts.value                         \\
121   & OutcomeClassTitle                       & outcomeMeasures.classes.title                               \\
122   & OutcomeClassDenomUnits                  & outcomeMeasures.classes.denoms.units                        \\
123   & OutcomeClassDenomCountGroupId           & outcomeMeasures.classes.denoms.counts.groupId               \\
124   & OutcomeClassDenomCountValue             & outcomeMeasures.classes.denoms.counts.value                 \\
125   & OutcomeCategoryTitle                    & outcomeMeasures.classes.categories.title                    \\
126   & OutcomeMeasurementGroupId               & outcomeMeasures.classes.categories.measurements.groupId     \\
127   & OutcomeMeasurementValue                 & outcomeMeasures.classes.categories.measurements.value       \\
128   & OutcomeMeasurementSpread                & outcomeMeasures.classes.categories.measurements.spread      \\
129   & OutcomeMeasurementLowerLimit            & outcomeMeasures.classes.categories.measurements.lowerLimit  \\
130   & OutcomeMeasurementUpperLimit            & outcomeMeasures.classes.categories.measurements.upperLimit  \\
131   & OutcomeMeasurementComment               & outcomeMeasures.classes.categories.measurements.comment     \\
132   & OutcomeAnalysisParamType                & outcomeMeasures.analyses.paramType                          \\
133   & OutcomeAnalysisParamValue               & outcomeMeasures.analyses.paramValue                         \\
134   & OutcomeAnalysisDispersionType           & outcomeMeasures.analyses.dispersionType                     \\
135   & OutcomeAnalysisDispersionValue          & outcomeMeasures.analyses.dispersionValue                    \\
136   & OutcomeAnalysisStatisticalMethod        & outcomeMeasures.analyses.statisticalMethod                  \\
137   & OutcomeAnalysisStatisticalComment       & outcomeMeasures.analyses.statisticalComment                 \\
138   & OutcomeAnalysisPValue                   & outcomeMeasures.analyses.pValue                             \\
139   & OutcomeAnalysisPValueComment            & outcomeMeasures.analyses.pValueComment                      \\
140   & OutcomeAnalysisCINumSides               & outcomeMeasures.analyses.ciNumSides                         \\
141   & OutcomeAnalysisCIPctValue               & outcomeMeasures.analyses.ciPctValue                         \\
142   & OutcomeAnalysisCILowerLimit             & outcomeMeasures.analyses.ciLowerLimit                       \\
143   & OutcomeAnalysisCIUpperLimit             & outcomeMeasures.analyses.ciUpperLimit                       \\
144   & OutcomeAnalysisCILowerLimitComment      & outcomeMeasures.analyses.ciLowerLimitComment                \\
145   & OutcomeAnalysisCIUpperLimitComment      & outcomeMeasures.analyses.ciUpperLimitComment                \\
146   & OutcomeAnalysisEstimateComment          & outcomeMeasures.analyses.estimateComment                    \\
147   & OutcomeAnalysisTestedNonInferiority     & outcomeMeasures.analyses.testedNonInferiority               \\
148   & OutcomeAnalysisNonInferiorityType       & outcomeMeasures.analyses.nonInferiorityType                 \\
149   & OutcomeAnalysisNonInferiorityComment    & outcomeMeasures.analyses.nonInferiorityComment              \\
150   & OutcomeAnalysisOtherAnalysisDescription & outcomeMeasures.analyses.otherAnalysisDescription           \\
151   & OutcomeAnalysisGroupDescription         & outcomeMeasures.analyses.groupDescription                   \\
152   & OutcomeAnalysisGroupId                  & outcomeMeasures.analyses.groupIds                           \\
[1ex]\multicolumn{3}{@{}l}{\textbf{resultsSection.adverseEventsModule}}\\[1ex]
153   & EventsFrequencyThreshold                & frequencyThreshold                                            \\
154   & EventsTimeFrame                         & timeFrame                                                     \\
155   & EventsDescription                       & description                                                   \\
156   & EventsAllCauseMortalityComment          & allCauseMortalityComment                                      \\
157   & EventGroupId                            & eventGroups.id                                                \\
158   & EventGroupTitle                         & eventGroups.title                                             \\
159   & EventGroupDescription                   & eventGroups.description                                       \\
160   & EventGroupDeathsNumAffected             & eventGroups.deathsNumAffected                                 \\
161   & EventGroupDeathsNumAtRisk               & eventGroups.deathsNumAtRisk                                   \\
162   & EventGroupSeriousNumAffected            & eventGroups.seriousNumAffected                                \\
163   & EventGroupSeriousNumAtRisk              & eventGroups.seriousNumAtRisk                                  \\
164   & EventGroupOtherNumAffected              & eventGroups.otherNumAffected                                  \\
165   & EventGroupOtherNumAtRisk                & eventGroups.otherNumAtRisk                                    \\
166   & SeriousEventTerm                        & seriousEvents.term                                            \\
167   & SeriousEventOrganSystem                 & seriousEvents.organSystem                                     \\
168   & SeriousEventSourceVocabulary            & seriousEvents.sourceVocabulary                                \\
169   & SeriousEventAssessmentType              & seriousEvents.assessmentType                                  \\
170   & SeriousEventNotes                       & seriousEvents.notes                                           \\
171   & SeriousEventStatsGroupId                & seriousEvents.stats.groupId                                   \\
172   & SeriousEventStatsNumEvents              & seriousEvents.stats.numEvents                                 \\
173   & SeriousEventStatsNumAffected            & seriousEvents.stats.numAffected                               \\
174   & SeriousEventStatsNumAtRisk              & seriousEvents.stats.numAtRisk                                 \\
175   & OtherEventTerm                          & otherEvents.term                                              \\
176   & OtherEventOrganSystem                   & otherEvents.organSystem                                       \\
177   & OtherEventSourceVocabulary              & otherEvents.sourceVocabulary                                  \\
178   & OtherEventAssessmentType                & otherEvents.assessmentType                                    \\
179   & OtherEventNotes                         & otherEvents.notes                                             \\
180   & OtherEventStatsGroupId                  & otherEvents.stats.groupId                                     \\
181   & OtherEventStatsNumEvents                & otherEvents.stats.numEvents                                   \\
182   & OtherEventStatsNumAffected              & otherEvents.stats.numAffected                                 \\
183   & OtherEventStatsNumAtRisk                & otherEvents.stats.numAtRisk                                   \\
[1ex]\multicolumn{3}{@{}l}{\textbf{resultsSection.moreInfoModule}}\\[1ex]
184   & LimitationsAndCaveatsDescription        & limitationsAndCaveats.description                                  \\
185   & AgreementPISponsorEmployee              & certainAgreement.piSponsorEmployee                                 \\
186   & AgreementRestrictionType                & certainAgreement.restrictionType                                   \\
187   & AgreementRestrictiveAgreement           & certainAgreement.restrictiveAgreement                              \\
188   & AgreementOtherDetails                   & certainAgreement.otherDetails                                      \\
189   & PointOfContactTitle                     & pointOfContact.title                                               \\
190   & PointOfContactOrganization              & pointOfContact.organization                                        \\
191   & PointOfContactEMail                     & pointOfContact.email                                               \\
192   & PointOfContactPhone                     & pointOfContact.phone                                               \\
193   & PointOfContactPhoneExt                  & pointOfContact.phoneExt                                            \\
[1ex]\multicolumn{3}{@{}l}{\textbf{derivedSection.conditionBrowseModule}}\\[1ex]
194   & ConditionMeshId                         & meshes.id                                                   \\
195   & ConditionMeshTerm                       & meshes.term                                                 \\
196   & ConditionAncestorId                     & ancestors.id                                                \\
197   & ConditionAncestorTerm                   & ancestors.term                                              \\
198   & ConditionBrowseLeafId                   & browseLeaves.id                                             \\
199   & ConditionBrowseLeafName                 & browseLeaves.name                                           \\
200   & ConditionBrowseLeafRelevance            & browseLeaves.relevance                                      \\
201   & ConditionBrowseBranchAbbrev             & browseBranches.abbrev                                       \\
202   & ConditionBrowseBranchName               & browseBranches.name                                         \\
[1ex]\multicolumn{3}{@{}l}{\textbf{derivedSection.interventionBrowseModule}}\\[1ex]
203   & InterventionMeshId                      & meshes.id                                                \\
204   & InterventionMeshTerm                    & meshes.term                                              \\
205   & InterventionAncestorId                  & ancestors.id                                             \\
206   & InterventionAncestorTerm                & ancestors.term                                           \\
207   & InterventionBrowseLeafId                & browseLeaves.id                                          \\
208   & InterventionBrowseLeafName              & browseLeaves.name                                        \\
209   & InterventionBrowseLeafRelevance         & browseLeaves.relevance                                   \\
210   & InterventionBrowseBranchAbbrev          & browseBranches.abbrev                                    \\
211   & InterventionBrowseBranchName            & browseBranches.name                                      \\
\end{longtable}

%% file: tables/facts-judge-prompts.tex
\begin{table*}[h]
\centering
\footnotesize
\resizebox{\textwidth}{!}{
\begin{tabular}{@{}llrrrrrr@{}}
\toprule
\textbf{Judge Model} & \textbf{Prompt Template} &
\textbf{Macro-F1} & \textbf{Acc.} &
\textbf{FPR} & \textbf{FNR} &
\textbf{F1 (+)} & \textbf{F1 ($-$)} \\
\midrule
\multirow{7}{*}{Claude 3.5 Sonnet}
& Span-level                  & 68.85 & 77.83 & 20.97 & 22.38 & 85.58 & 52.13 \\
& \textbf{Implicit span-level} & \textbf{70.24} & 83.50 & 45.16 & 11.34 & 90.10 & 50.37 \\
& Response-level              & 61.88 & 83.25 & 72.58 & 6.69  & 90.42 & 33.33 \\
& JSON                        & 56.04 & 64.78 & 33.87 & 35.47 & 75.64 & 36.44 \\
& JSON (alt)                  & 55.37 & 66.75 & 46.77 & 30.81 & 77.91 & 32.84 \\
& JSON w. double-check        & 49.50 & 54.68 & 25.81 & 48.84 & 65.67 & 33.33 \\
& SimpleQA template           & 55.39 & 85.22 & 88.71 & 1.45  & 91.87 & 18.92 \\
\midrule
\multirow{7}{*}{Gemini 1.5 Pro}
& Span-level                  & 55.84 & 79.31 & 79.03 & 10.17 & 88.03 & 23.64 \\
& Implicit span-level         & 56.66 & 85.47 & 87.10 & 1.45  & 91.99 & 21.33 \\
& Response-level              & 48.82 & 82.02 & 95.16 & 4.07  & 90.04 & 7.59  \\
& \textbf{JSON}               & \textbf{71.47} & 86.95 & 56.45 & 5.23  & 92.48 & 50.47 \\
& JSON (alt)                  & 66.03 & 85.96 & 69.35 & 4.07  & 92.05 & 40.00 \\
& JSON w. double-check        & 64.89 & 76.35 & 37.10 & 21.22 & 84.95 & 44.83 \\
& SimpleQA template           & 51.54 & 84.73 & 93.55 & 1.16  & 91.64 & 11.43 \\
\midrule
\multirow{7}{*}{GPT-4o}
& Span-level                  & 63.08 & 81.53 & 64.52 & 10.17 & 89.18 & 36.97 \\
& Implicit span-level         & 55.43 & 83.99 & 87.10 & 3.20  & 91.11 & 19.75 \\
& Response-level              & 51.54 & 84.73 & 93.55 & 1.16  & 91.64 & 11.43 \\
& \textbf{JSON}               & \textbf{69.68} & 80.54 & 32.26 & 17.15 & 87.83 & 51.53 \\
& JSON (alt)                  & 66.78 & 82.02 & 53.23 & 11.63 & 89.28 & 44.27 \\
& JSON w. double-check        & 57.62 & 64.04 & 17.74 & 39.24 & 74.11 & 41.13 \\
& SimpleQA template           & 47.04 & 83.74 & 98.39 & 1.45  & 91.13 & 2.94 \\
\bottomrule
\end{tabular}
}
\caption{Evaluation of judge models and prompt templates, reproduced from the \textbf{FACTS Grounding Benchmark} paper (Table~2).}
\label{tab:facts-judge-prompts}
\end{table*}